\RequirePackage{fix-cm}
\documentclass[smallextended]{svjour3}       
\smartqed  
\usepackage{graphicx}
\usepackage{subfig}
\usepackage{graphicx}
\usepackage{multirow}
\usepackage{booktabs, makecell, rotating}
\usepackage{tikz}
\usepackage{amsmath,amssymb}
\usepackage[numbers]{natbib}
\usepackage{url}

\usepackage{breakurl}
\usepackage[breaklinks]{hyperref}

\usepackage{xcolor}
\usepackage{color, soul}
\newcommand{\todo}[1]{\textcolor{red}{\hl{[#1]}}}

\newcommand{\resp}[1]{\textcolor{purple}{[#1]}}
\newcommand{\nop}[1]{}
 
\newcommand{\figwidthtwo}{0.47} 

\begin{document}
\begin{sloppy}
\title{
\bf Data Pricing in Machine Learning Pipelines
}


\author{Zicun Cong         \and
        Xuan Luo \and
        Pei Jian \and
        Feida Zhu \and
        Yong Zhang
}


\institute{
Zicun Cong \at 
Simon Fraser University, Burnaby, Canada \\
\email{zicun\_cong@cs.sfu.ca}
\and
Xuan Luo \at
Simon Fraser University, Burnaby, Canada\\
\email{xuan\_luo@cs.sfu.ca}
\and
Jian Pei \at
Simon Fraser University, Burnaby, Canada\\
\email{jpei@cs.sfu.ca}
\and
Feida Zhu \at
Singapore Management University, Singapore\\
\email{fdzhu@smu.edu.sg}
\and
Yong Zhang \at
Huawei Technologies Canada, Burnaby, Canada\\
\email{yong.zhang3@huawei.com}
}

\date{Received: date / Accepted: date}

\maketitle

\begin{abstract}
Machine learning is disruptive.  At the same time, machine learning can only succeed by collaboration among many parties in multiple steps naturally as pipelines in an eco-system, such as collecting data for possible machine learning applications, collaboratively training models by multiple parties and delivering machine learning services to end users.  Data is critical and penetrating in the whole machine learning pipelines.  As machine learning pipelines involve many parties and, in order to be successful, have to form a constructive and dynamic eco-system, marketplaces and data pricing are fundamental in connecting and facilitating those many parties.  In this article, we survey the principles and the latest research development of data pricing in machine learning pipelines. We start with a brief review of data marketplaces and pricing desiderata.  Then, we focus on pricing in three important steps in machine learning pipelines.  To understand pricing in the step of training data collection, we review pricing raw data sets and data labels.  We also investigate pricing in the step of collaborative training of machine learning models, and overview pricing machine learning models for end users in the step of machine learning deployment. We also discuss a series of possible future directions.

\end{abstract}

\section{Introduction}
\label{sec:introduction}
\nop{
\todo{Yong: After reading the whole paper, I feel we mainly focus on supervised machine learning pipeline. Is this true? If this is true, should we make it clear about this. Otherwise, should we discuss the difference between supervised and unsupervised machine learning in data pricing?
Some different points I thought are:
1) There is no labeling in unsupervised learning.
2) The quality of the model from unsupervised learning may be evaluated based on various downstream tasks. Will the existing pricing algorithms still work in this kind of scenario?
3) Similar issues may also exist in pricing in collaborative training.}
\resp{Zicun's response: Our survey does not only focus on supervised machine learning pipelines. Many of the discussed pricing models can be used in machine learning pipelines of both unsupervised and supervised machine learning models. First, the existing pricing models of raw data sets are agnostic to the down-stream applications of the data sets. Second, for pricing models in collaborative training, participants are rewarded according to the contributions of their data sets to the utility of the jointly trained machine learning model. To adapt existing methods to scenarios where unsupervised machine learning models are trained, the key challenge is to develop a utility function that participants all agree on. Therefore, our survey covers pricing techniques for both supervised and unsupervised machine learning models. For more details, please refer to the end of Sections~\ref{sec:price_raw} and~\ref{sec:price_collaborative}.}
}
The disruptive success of machine learning in many applications has led to an explosion in demand~\citep{DBLP:conf/uss/TramerZJRR16, agarwal2019marketplace}. Recent research predicts that the global machine learning market is expected to reach 20.83 billion dollars in 2024~\citep{forbes}. To succeed in building a machine learning application, one party is far from enough.  Many parties have to collaborate in one way or another.  For example, one party may have to acquire raw data and data labeling services from some other parties to construct training data, multiple parties may need to collaborate in building a machine learning model, and one party may want to use some other parties' models to solve its business problems. Machine learning applications are indeed pipelines connecting many parties. 

Data is critical for machine learning.  Machine learning models, especially deep models, rely on large amounts of data for training and testing.  Deploying machine learning services also needs data -- machine learning models consume users' data as input, and return insights and recommend possible actions. Maintaining and updating machine learning models still need data.  The importance of data for machine learning cannot be over emphasized.  Data penetrates the whole machine learning pipelines.

Obtaining data for machine learning is far from easy~\citep{liu2020dealer}. For a party that wants to build a machine learning model, the challenges come from multiple aspects.  First, within the party, in order to develop a training data set, more often than not it is costly to collect data, create proper labels and ensure data quality. Second, the party may realize that it does not have the necessary data to train the target model.  Thus, the party may have to explore external sources for the data needed.  This involves acquiring external data.  Last, to build or strengthen business edges, the party may want to provide machine learning services to other parties.  Then, the party has to exchange data with other parties, such as accessing data from end users and providing end users model output.

Connecting many parties in an eco-system in scale requires a general and principled mechanism. As data and models are essential in machine learning pipelines and data and model exchanges are the most fundamental interactions among different parties, data and model marketplaces become a natural choice for machine learning pipelines and eco-systems, and pricing becomes the core mechanism in machine learning pipelines.

In response to the massive and diversified demands for various data, data products become valuable assets for purchase and sale. Here, \emph{data products} refer to data sets as products and information services derived from data sets~\citep{pei2020survey}.
Data commoditization motivates data owners to share their data products in exchange of rewards and thus helps data buyers to access data products of high quality and large quantities. 

To enable tradings between data owners and data buyers, data must be priced. Pricing data, however, is far from trivial. \citet{agarwal2019marketplace} summarize five properties making data a unique asset. First, data can be replicated at zero marginal cost. Second, the value of data is inherently combinatorial. Third, the value of data varies widely among different buyers. Last, the usefulness of data lies in the value of information derived from it, which is difficult to verify a priori. Due to those properties, pricing models for physical goods cannot be directly applied  or straightforwardly extended to data products, and thus new principles, theories, and methods need to be developed.

\begin{figure}[t]
    \centering
    \includegraphics[width=0.8\linewidth, page=12]{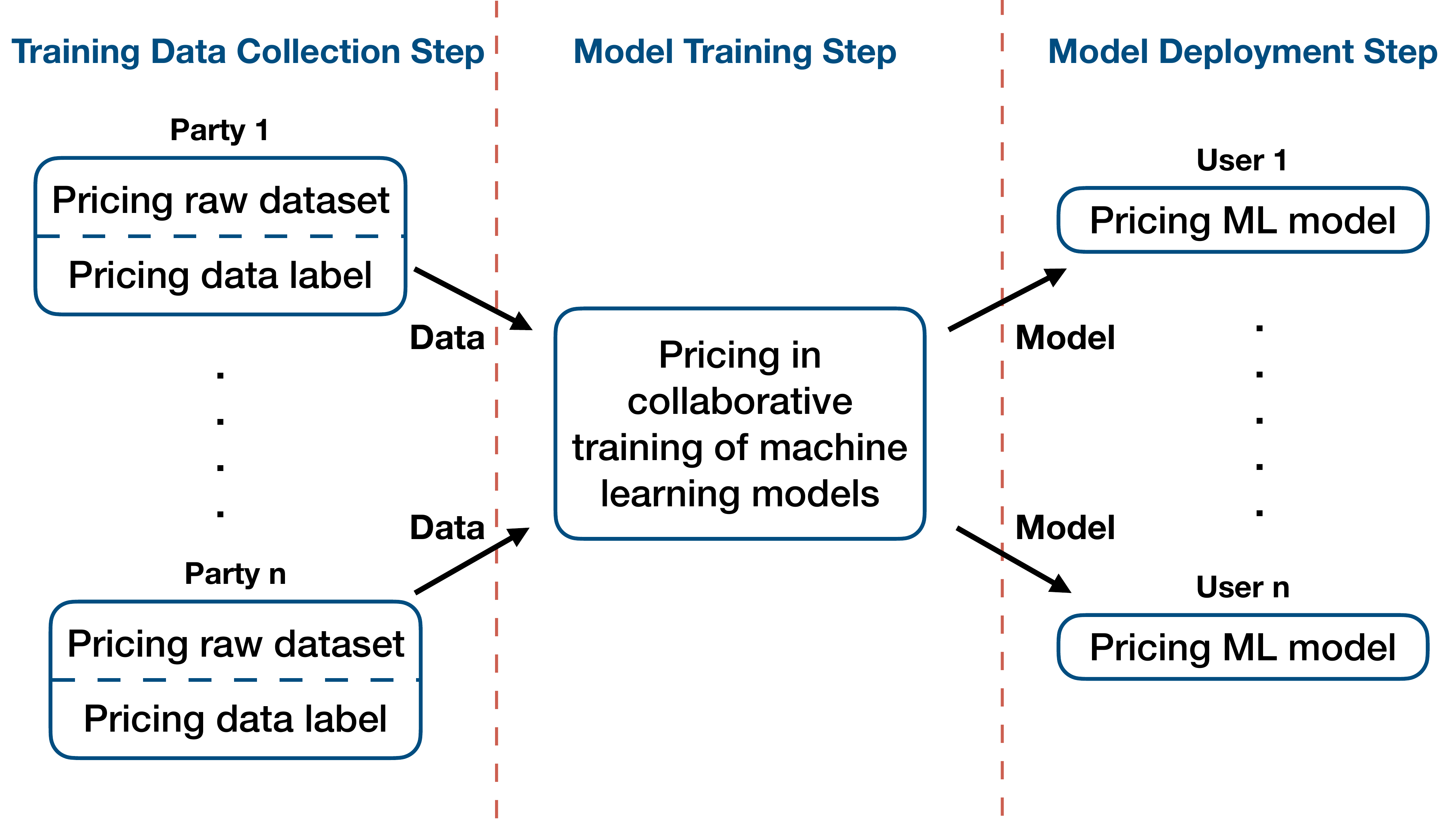}
    \caption{Steps and pricing tasks in machine learning pipelines}
    \label{fig:pipeline}
\end{figure}

Based on an extensive survey on existing research, we identify and focus on three steps in data and model supply tasks in the manufacturing pipeline of machine learning models~\citep{fung2019brokered}. The steps and their corresponding tasks are illustrated in Figure~\ref{fig:pipeline}. In the step of training data collection, raw data is collected and the associated labels are annotated. We review the research on pricing for raw data sets and data labels.  In the step of collaborative training of machine learning models, we investigate how to price different participants' contributions through their data.  In the model deployment step, we overview pricing machine learning models for end users.  We focus on the four pricing tasks in machine learning pipelines as follows.

\begin{itemize}

\item \emph{Pricing raw data sets}. To build a machine learning model, the first step is collecting training data.  Monetizing and trading raw data sets provide people with a convenient and efficient way to acquire a large amount of training data. A key challenge in pricing raw data sets is how to set the price reflecting the usefulness of a data set.  Moreover, pricing models may be optimized towards different objectives, such as revenue maximization, arbitrage-freeness, and truthfulness. Achieving those optimization goals introduces additional challenges in the design and implementation of pricing models.

\item \emph{Pricing data labels.} In the training data collection step, in addition to collecting raw data,  obtaining data labels is critical. Crowdsourcing is a popular way for this purpose~\citep{vaughan2017making}. 
Unfortunately, spammers may commit no efforts in their assigned tasks and produce random answers, which leads to data sets of poor quality. 
Thus, a key challenge in pricing data labels is how to estimate label accuracy and compensate crowdworkers correspondingly, such that they are motivated and driven to invest high efforts and report accurate data labels~\citep{vaughan2017making}. This task only appears in machine learning pipelines where supervised machine learning models are produced.

\item \emph{Revenue allocation in collaborative machine learning.}
Collaborative machine learning is an emerging paradigm, where multiple data owners collaboratively train machine learning models on their aggregate data, and share the revenues of using/selling these models. Data sets from different owners may have different contributions to the learned models. Evenly distributing the revenues is not fair to data owners, particularly for those who contribute more valuable data, and thus may discourage future collaborations. To this end, a key challenge is how to fairly reward data owners' contributions. 
\nop{\todo{Yong: Should we consider the machine learning training cost in the revenue allocation in collaborative machine learning?}
\resp{Zicun's response: I add a study in Section~\ref{sec:price_collaborative}, where training costs in collaborative machine learning are considered.}
}

\item \emph{Pricing machine learning models.} 
Machine learning as a service (MLaaS)~\citep{DBLP:conf/uss/TramerZJRR16, chen2020frugalml} is a rapidly growing industry. 
Customers may purchase well-trained machine learning models or build models on top of those well-trained rather than building models from scratch by themselves. For example, one may use Google prediction API to classify an image for only \$0.0015~\citep{chen2020frugalml}. While machine learning models and raw data sets share a series of common ideas in pricing, the pricing models of raw data sets cannot be trivially adapted to price machine learning models. How to version machine learning models and avoid arbitrage among multiple versions is a key challenge in this task.

\end{itemize}

The four tasks are related to each other. They share some core ideas, that is, linking prices of data products to their utilities to customers. But as the tasks have different application scenarios and pricing goals, they are solved by orthogonal techniques.

The existing models in the first two tasks aim at pricing training data sets with absolute utility functions, that is, the utility of a data product only depends on the properties of the product. 
One important difference between the first two tasks is about the utility functions. The utility (e.g., accuracy) of data labels is very hard to compute due to the lack of ground-truth verifications. 
The third task evaluates the utility of a data set by its marginal contribution to a machine learning model. Thus, the utility of a data set also depends on the utility of other data sets used to jointly build the model. The existing methods in the last task also employ absolute utility functions. But as machine learning models and data sets have different properties, new pricing models are developed.

The four tasks are connected when machine learning models and data sets are priced in an end-to-end manner. On the one hand, the price of a machine learning model limits the budget of training data procurement and the revenue that can be split among data owners~\citep{agarwal2019marketplace}. 
On the other hand, the costs of data procurement and model training also influence the selling price of machine learning models, as they are part of the manufacturing cost~\citep{liu2020dealer}.
Figure~\ref{fig:pipeline} shows the connections of the four tasks.

There are some previous surveys related to data pricing~\citep{liang2018survey, 10.1007/978-3-319-69191-6_4, ZhangBletran20}. This article covers a substantially deeper and more focused scope than those. In this article, we try to present a comprehensive survey on data pricing in machine learning pipelines.
Very recently, \citet{pei2020survey} presents a survey connecting economics, digital product pricing, and data product pricing. 
He identifies a series of desirable properties in data pricing and reviews the  techniques achieving those properties. But~\citet{pei2020survey} does not focus on machine learning pipeline and does not cover the studies of pricing data labels.

The rest of this survey is organized as follows. 
Section~\ref{sec:basic} reviews basic concepts and essential principles in data product pricing. 
Section~\ref{sec:price_raw} reviews pricing raw data sets. Pricing data labels is discussed in Section~\ref{sec:price_label}. In Section~\ref{sec:price_collaborative}, we review the recent progress in revenue allocation in collaborative machine learning. Section~\ref{sec:price_model} is about how to price machine learning models. We conclude this survey and discuss some future directions in Section~\ref{sec:conclusion}.

\section{Data Marketplaces and Pricing}
\label{sec:basic}

In this section, we briefly review data marketplaces and pricing in general. We first discuss the basic structures of data marketplaces. Then, we discuss some major pricing strategies in general.  Third, we discuss different types of data markets in terms of competition and dominance.  Last, we discuss the desiderata of data pricing. 

\subsection{Data Marketplaces}


A data marketplace is a platform that allows people to buy and sell data products~\citep{schomm2013marketplaces}. 
Some examples of data marketplaces include Dawex~\citep{DAWEX}, Snowflake data marketplace~\citep{SNOW}, and BDEX~\citep{BDEX}. 
\citet{muschalle2012pricing} identify seven categories of participants in data marketplaces, namely analysts, application vendors, data processing algorithm developers, data providers, consultants, licensing and certification entities, and data market owners.

\begin{figure}[t]
\centering
\subfloat[General data marketplace.]{
\includegraphics[page=3, width=0.8\linewidth, trim=0 600 450 0, clip,]{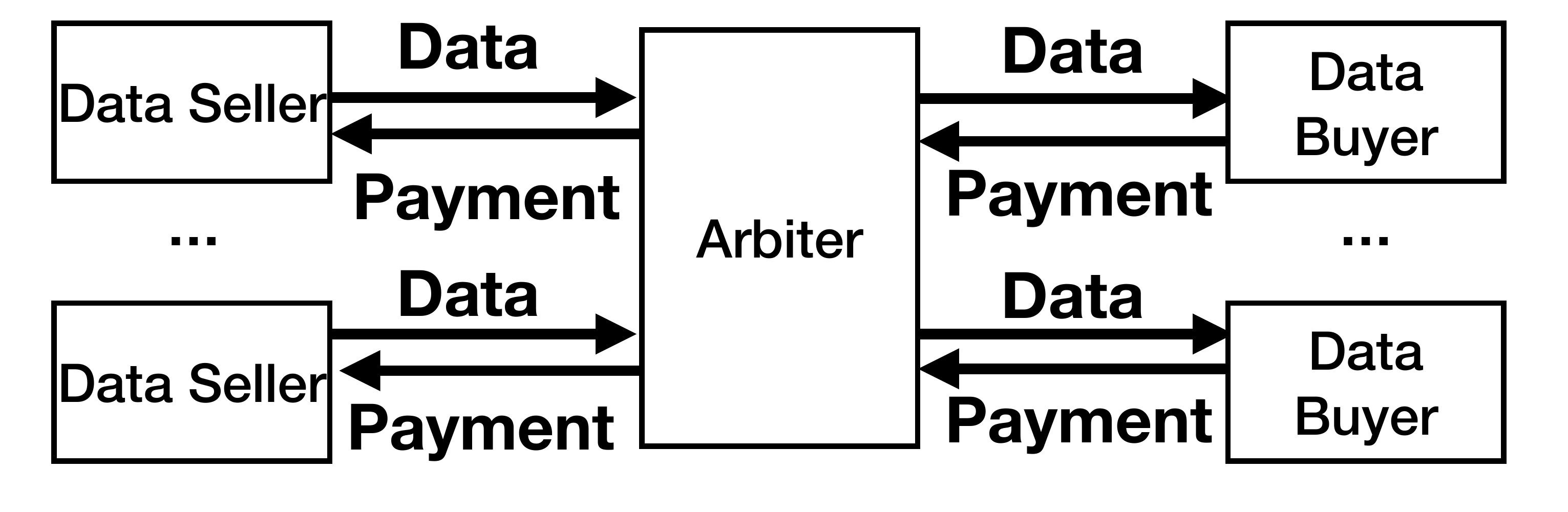}
\label{fig:concise_sr_ais}
}
\qquad
\subfloat[Sell-side marketplace.]{
\includegraphics[page=8, width=\figwidthtwo\linewidth, trim=0 600 1000 0, clip]{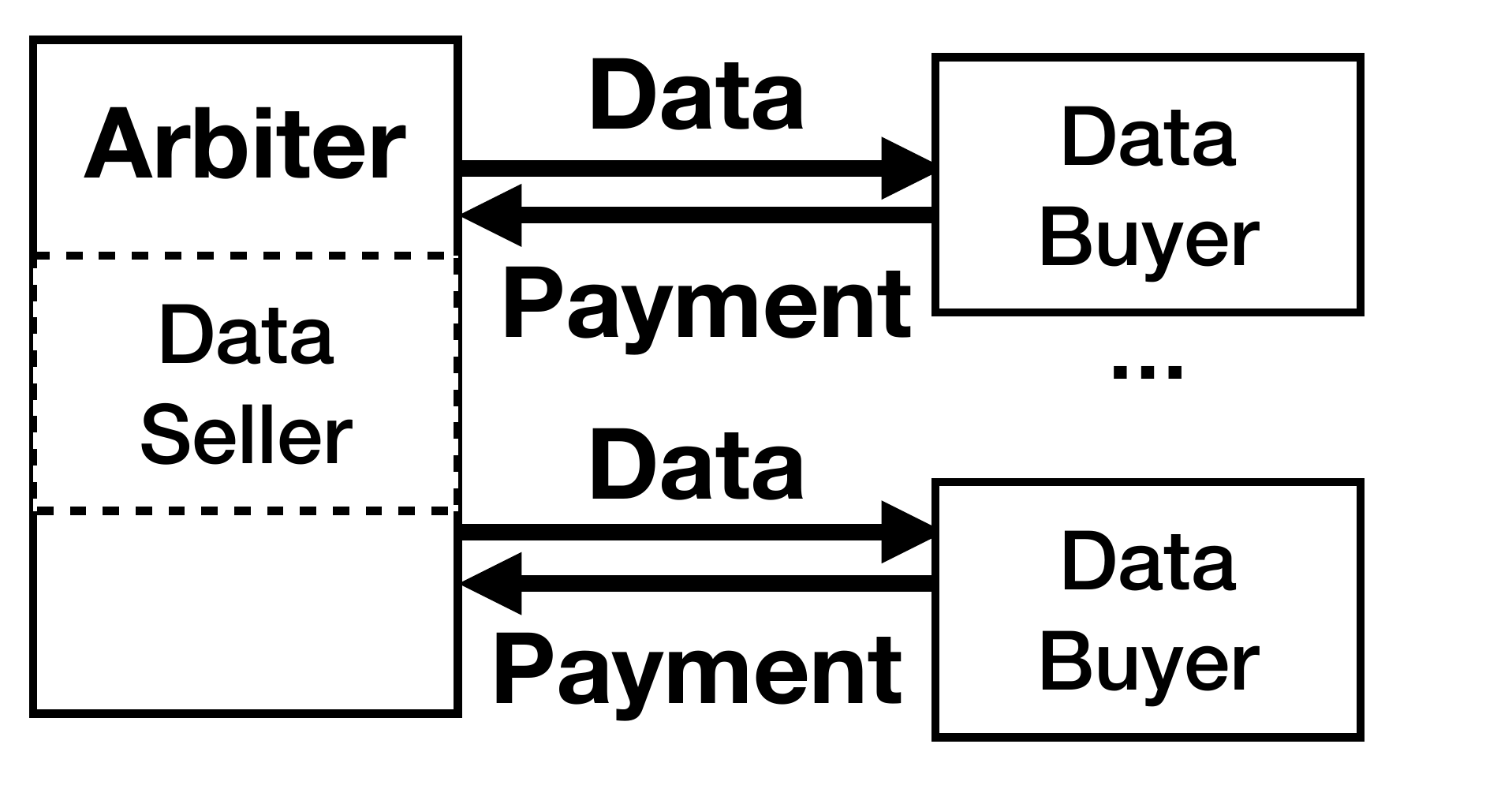}
\label{fig:sell_side}
}
\subfloat[Buy-side marketplace.]{
\includegraphics[page=7, width=\figwidthtwo\linewidth, trim=0 600 1000 0, clip]{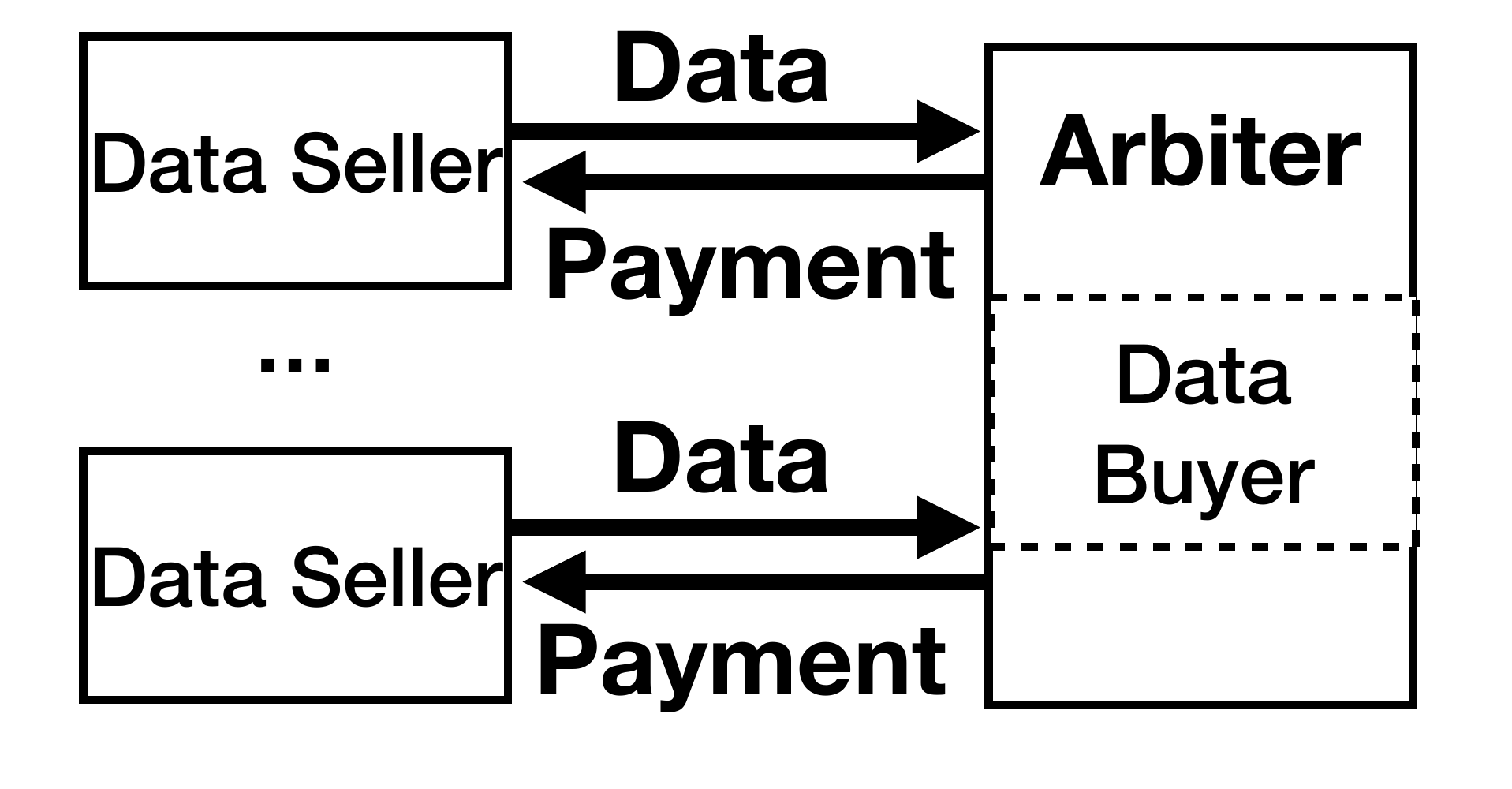}
\label{fig:buy_side}
}
\caption{Architectures of data marketplace.}
\label{fig:marketplace}
\end{figure}

Figure~\ref{fig:concise_sr_ais} shows the conceptual architecture of data marketplaces.  A data marketplace mainly consists of three major entities, namely data sellers, an arbiter (also known as data vendor~\citep{schomm2013marketplaces} and data broker~\citep{DBLP:conf/kdd/NiuZWTGC18}), and data buyers. Data sellers own data products and are willing to share those products with the arbiter in exchange for rewards. Data buyers want to obtain data products to solve their problems. The function of an arbiter is to facilitate transactions between data sellers and data buyers. The arbiter collects data products from data sellers and sells them to data buyers. After collecting the payments from buyers, the arbiter distributes the payments to data sellers. In general, arbiters are modeled as non-profit participants in data marketplaces.

Some studies simplify the architecture of data marketplaces to sell-side marketplaces and buy-side marketplaces. A sell-side marketplace~\citep{ZhangBletran20}, as shown in Figure~\ref{fig:sell_side}, has a single data provider and multiple data buyers. In a sell-side marketplace, the arbiter is operated by a monopoly data seller to sell the single seller's data products. 
In literature, sell-side marketplaces are considered by pricing models of both general data sets~\citep{heckman2015pricing} and specific types of data products, such as XML documents~\citep{DBLP:conf/dexa/TangASB14} and data queries on a relational database~\citep{DBLP:conf/sigmod/DeepK17}.


A buy-side marketplace~\citep{ZhangBletran20}, as shown in Figure~\ref{fig:buy_side}, has multiple data providers and a single consumer/data buyer. In a buy-side marketplace, the arbiter is operated by the single data buyer for purchasing data products from providers. Buy-side marketplaces are considered in many existing studies~\citep{ghorbani2019data, de2016incentives, jia2019efficient}. For instance, \citet{de2016incentives} study a buy-side marketplace, where a single consumer pays crowdsource workers for labeling the single buyer's data set.


\subsection{Pricing Strategies}

Many pricing strategies have been developed in pricing theory. Cost-based pricing, customer value-based pricing, and competition-based pricing are three important categories~\citep{de2017pricing}.

Cost-based pricing considers that the price of a product is determined by adding a specific amount of markup to the cost. This strategy is adopted in personal data pricing, where the cost is the total privacy compensation to data owners~\citep{DBLP:conf/kdd/NiuZWTGC18}. A disadvantage of the cost-based pricing strategy is that it only considers internal factors in determining the selling price. External factors, such as competition and demands, are not included~\citep{luong2016data}. 

Customer value-based pricing determines the price of a product primarily based on how much the target customers believe a product is worth~\citep{de2017pricing}. To apply customer value-based pricing, a seller needs to estimate customers' demands for a product through their willingness and affordability~\citep{luong2016data}. Customer value-based pricing is the most popularly used strategy for data pricing.

Competition-based pricing determines the price of a product strategically based on competitors' price levels and behavior expectations~\citep{de2017pricing}. Game theory provides a powerful tool to implement the strategy. In a non-cooperative game, every seller is selfish and sets the price that maximizes the seller's profit independently~\citep{luong2016data}. The competition result, that is the asking price of each seller, is the Nash equilibrium~\citep{Nash50}. 

There are some other major pricing strategies in literature~\citep{oro2041, 2135, Brennan13, 83820, Irvin79}, such as operation-oriented pricing, revenue-oriented pricing, and relationship-oriented pricing. The remarkably rich body of studies in economics and marketing research on pricing tactics is far beyond the scope and capacity of this survey.

\subsection{Four Types of Data Markets}
Similar to physical goods, the prices of data products are also influenced by the dominance and diversity of supplies and demands in the market.

\citet{10.1007/978-3-319-69191-6_4} identify four types of data markets.
First, in a \emph{monopoly}, a supplier holds enough market power to set prices to maximize profits. Second, in an \emph{oligopoly}, a small number of suppliers dominate the market.
Third, in \emph{strong competition markets}, individual suppliers do not have enough market powers to set profit-maximizing prices, and prices tend to align with marginal costs. Last, in a \emph{monopsony}, a single buyer controls the market as the only consumer of products provided by sellers. 

Most studies assume explicitly or implicitly a monopoly (monopsony) market structure where the data seller (data buyer) does not care about competing with others. Data pricing in oligopoly market is considered by~\citet{ balasubramanian2015pricing}. \citet{jiang2015economics} study a perfect competition market where participants can directly trade with each other.

\subsection{Desiderata of Data Pricing}
\label{sec:desiderata}

There are some desiderata perferred by most pricing models. In this section, we briefly review the six desiderata suggested by \citet{pei2020survey}.
In addition, we complement the existing study by an important desideratum, effort elicitation.

\paragraph{Truthfulness} Truthfulness is an important economic property of robust markets~\citep{zhou2009trust}. In a truthful market, all participants are selfish and only offer prices that maximize their utility values. Participants may have their own valuations on the same product, but a truthful market guarantees that for each participant, offering the real valuation is an individual's best strategy. In other words, no participants will lie about their valuations. Truthfulness simplifies all participants' strategies and ensures basic market fairness~\citep{10.14778/3407790.3407800}. 

Reverse auction is a common tool to implement truthful data markets.
In a reverse auction, $N$ sellers $D = \{s_1, \ldots, s_N\}$ compete for a buyer's deal by submitting their asking prices $\{b_1, \ldots, b_N\}$. An auction mechanism takes as input the submitted bids, selects a subset of sellers as winners, and determines the payment $p_i$ to each winner $s_i$, where $p_i \geq b_i$. 
In a truthful reverse auction, the best strategy (dominant strategy) for a seller $s_i$ to maximize the expected utility is submitting the individual's real valuation, no matter what others submit. 

In his seminal paper on optimal mechanism design,~\citet{myerson1981optimal} shows that a sealed-bid reverse auction mechanism is truthful if and only if (1) The selection rule is monotone, that is, if a seller $s_i$ wins the auction by bidding $b_i$, it also wins by bidding $b'_i \leq b_i$; and (2) Each winner is paid the critical value, that is, seller $s_i$ would not win the auction if $s_i$ bids higher than this value. 

\paragraph{Revenue Maximization}

Revenue maximization is a strategy to increase a seller's customer base by having low prices. This strategy is widely adopted by sellers in an emerging market to build market share and reputations. 
For traditional physical goods, the curves of marginal cost are U-shaped with respect to manufacturing level. The revenue of a seller is maximized when the manufacturing level is set such that the marginal revenue is zero~\citep{BurkettJohnP2006MOEa}. Since data products can be re-produced at almost zero costs~\citep{agarwal2019marketplace}, the revenue maximization techniques for data products and physical products are quite different~\citep{pei2020survey}.
 
\paragraph{Fairness}

In some scenarios, sellers need to cooperatively participate in a transaction. A data market is fair to the contributors in a coalition if the revenue generated by the coalition is fairly divided among the sellers. 

Suppose a set of sellers $D=\{s_1, \ldots, s_N\}$ cooperatively participate in a transaction that leads to a payment $v$. \citet{shapley1953value} lays out four axioms for a fair allocation.

\begin{itemize}
    \item \textit{Balance}: The payment $v$ should be fully distributed to the sellers in $D$.
    \item \textit{Symmetry}: Sellers making the same contribution to the payment should be paid the same. For a set of sellers $S$ and two additional sellers $s$ and $s'$, if $S \cup \{s\}$ and $S \cup \{s'\}$ lead to the same payment, sellers $s$ and $s'$ should get the same payment.
    \item \textit{Zero element}: If a seller's data does not contribute to the payment of any coalitions,
    the seller
       should receive no payment. 
    \item \textit{Additivity}: If the data of a group of sellers can be used for two tasks $t_1$ and $t_2$ with payments $v_1$ and $v_2$, respectively, then the payment to solve both tasks $t_1 + t_2$ should be $v_1 +v_2$.
\end{itemize}

It is proved that Shapley value $\psi(s)$ is the unique allocation method that satisfies the four axioms, which is defined as the average marginal contribution of $s_i$ to all possible subsets of sellers $S \subseteq D \setminus \{s_i\}$
\begin{equation}
    \psi(s) = \frac{1}{N}\sum_{S \subseteq D \setminus \{s\}}\frac{\mathcal{U}(S \cup \{s\}) - \mathcal{U}(S)}{ {N-1 \choose |S|} },
    \label{eq:shapley_1}
\end{equation}
where $\mathcal{U}(\cdot)$ is the utility function~\citep{shapley1953value}.  For example, in the context of collaborative machine learning, $\mathcal{U}(S)$ is the performance score of the machine learning model trained on the data sets of $S$, such as precision.

Equation~\ref{eq:shapley_1} can be rewritten to
\begin{equation}
    \psi(s) = \frac{1}{N!}\sum_{\pi \in \prod(D)}(\mathcal{U}(P_s^{\pi} \cup \{s\} - \mathcal{U}(P_s^{\pi}))),
    \label{eq:shapley_2}
\end{equation}
where $\pi \in \prod(D)$ is a permutation of sellers and $P_s^{\pi}$ is the set of sellers that precede seller $s$ in $\pi$.

The fact that Shapley value uniquely possesses Shapley fairness, combined with its flexibility to support different utility functions, makes it a popular tool to implement fair data marketplaces.

\paragraph{Arbitrage-free Pricing}
Arbitrage is the activities that take advantage of price differences between multiple markets. In a data marketplace, a data seller may offer multiple versions of products. As a consequence, a critical concern is that a data buyer may circumvent the advertised price of a product through buying a bundle of cheaper ones, which negatively affects the seller's revenue. For example, consider a data seller selling noisy queries to the seller's database~\citep{DBLP:conf/kdd/NiuZWTGC18, pei2020survey}, and the seller perturbs each query answer independently with random noise. An answer with a variance of $5$ is sold at \$$5$ and with a variance of $1$ is sold at \$$50$. A data buyer wants to obtain an answer of variance $1$. The buyer can purchase the cheaper answer $5$ times and compute their average. Since the noises are added independently, the aggregated average has variance $1$. Thus the customer saves \$25 by arbitrage. A desirable pricing function should guarantee that no arbitrage is possible, in which case we call it arbitrage-free.

\paragraph{Privacy-preservation} Privacy protection during the transactions of data raises more and more concerns. In data marketplaces, the privacy of buyers, sellers, and involved third parties are highly vulnerable, and might be disclosed in many different ways~\citep{pei2020survey}. Many different solutions have been proposed for privacy protection in data markets~\citep{10.14778/3407790.3407800, DBLP:conf/icdt/LiLMS13, DBLP:journals/pvldb/HynesDYCS18}.
In this survey, we focus on the studies along the line of privacy compensation~\citep{DBLP:conf/icdt/LiLMS13, DBLP:conf/kdd/NiuZWTGC18, DBLP:conf/sigir/NgetCY17}, which investigate how to provide compensations for the privacy disclosure of data owners. 
For the purpose of privacy protection, sensitive data sets are usually traded with injected random noise~\citep{DBLP:conf/sigir/NgetCY17}. A data set with less random noise is more accurate, but may leak more privacy and thus more compensations should be made to the data owner.

\paragraph{Computational Efficiency}

The numbers of transactions, sellers and buyers may be huge in a data marketplace. Therefore, it is a fundamental requirement for a pricing model to compute prices efficiently with respect to a large number of goods and participants. 
Prices should be computed in polynomial time with respect to the number of participants~\citep{agarwal2019marketplace} or the number of data products~\citep{chen2019towards}.
In some application scenarios, however, it takes exponential time to compute the pricing functions with desirable properties, such as Shapley fairness~\citep{ghorbani2019data}, arbitrage-freeness~\citep{DBLP:conf/pods/KoutrisUBHS12}, and revenue maximization~\citep{chen2019towards}.
For example, \citet{DBLP:conf/pods/KoutrisUBHS12} show that computing arbitrage-free prices of join queries on a relational database is in general NP-hard. How to efficiently determine prices with desirable properties  presents technical challenges.

\paragraph{Effort Elicitation}

In addition to the above six desiderata, here we propose a new one, effort elicitation.

In a data marketplace, a data buyer may purchase training data labels via crowdsourcing. Crowdworkers are presented with unlabeled data instances (for instance, images) and are asked to provide labels (for instance, a binary label indicating whether or not the image contains pedestrains). A major challenge in label collection is to ensure that workers invest their efforts and provide accurate answers. A poorly designed pricing model may result in labels with very low quality~\citep{shah2016double}. For example, if each task has a fixed price, an obvious strategy that maximizes a worker's profit is to just provide arbitrary answers without even solving the tasks~\citep{vaughan2017making}. Many techniques have been developed to post-process noisy answers in order to improve their quality. However, when the inputs to these algorithms are highly erroneous, it is difficult to guarantee that the processed answers will be reliable enough for downstream machine learning tasks~\citep{shah2016double}. In order to avoid the troubles of ``garbage in, garbage out'', a desirable approach is to design proper rewards for crowdsourcing tasks that  incent workers to invest efforts and provide higher quality answers~\citep{vaughan2017making}.

\section{Pricing Raw Data Sets}
\label{sec:price_raw}

In this section, we review the existing studies focusing on pricing raw data sets. The existing studies consider four types of scenarios. 
The most traditional methods price data sets as indivisible units and do not consider supplier competitions.
The intrinsic properties of data sets, such as volumes, are factors determining prices. In the second scenario, how to price indivisible data sets in a competitive market is studied. In the third scenario, data consumers can purchase just a fraction of an entire data set, which is more flexible  to consumers but may have the issue of arbitrage. The last scenario addresses pricing personal data by privacy compensation.

\subsection{Pricing General Data}
Machine learning and statistical models are vulnerable to poor quality training data, thus
high quality data is valuable to data buyers~\citep{vaughan2017making}. Pricing data sets based on quality becomes a natural choice. 

\citet{heckman2015pricing} identify a list of factors to assess the quality of a data set, such as age of data, accuracy of data, and volume of data. A linear model is proposed to set the price of a data set as
\begin{equation*}
    \text{price }=\text{Fixed cost} + \sum_{i} w_i \cdot \text{factor}_i.
\end{equation*}
Estimating the model parameters $w_i$ is a difficult task, as many data sets may not have public prices associated with them. A more comprehensive list of quality criteria is proposed in~\citep{DBLP:conf/aciids/StahlV16}.

\citet{DBLP:journals/candie/YuZ17} study the problem of trading multiple versions of a data set, constructed by different data quality factors. They assume customers' demands and maximum acceptable prices of different versions are public. 
A bi-level programming model is established to address the problem. At the first level, the data seller determines versions and their prices to maximize the total revenue. At the second level, a group of buyers select data products to maximize their utilities. Solving the bi-level programming model is NP-hard. \citet{DBLP:journals/candie/YuZ17} propose a heuristic genetic algorithm to approach it numerically.

\subsection{Pricing Crowdsensing Data}

Crowdsensing is a powerful tool to quickly and cheaply obtain vast amounts of training data for machine learning models~\citep{vaughan2017making, DBLP:journals/comsur/ZhangYSLTXM16}. In a crowdsensing marketplace, a task requester initiates a data collection task and compensates participating workers according to their reported  costs. As workers may exaggerate their costs, pricing models should incentivize workers to truthfully reveal their costs.


\citet{yang2012crowdsourcing} design a reverse auction mechanism for mobile sensing data, that is truthful, individually rational, and profitable. A pricing model is truthful if all sellers truthfully report their data collection costs.  A model is individually rational if all sellers have non-negative net profits, and profitable if the data buyer has non-negative net profits. The authors assume that a buyer has a set $\Gamma = \{\tau_1, \ldots, \tau_n\}$ of sensing tasks, where each task $\tau_i$ has a value $v_i$ to the buyer. Each seller $s_i$ chooses a subset of tasks $\Gamma_i \subseteq \Gamma$ and has a private cost $c_i$ for performing the tasks. Seller $s_i$ decides a price $b_i$ for the sensed data and submits the task-bid pair $(\Gamma_i, b_i)$ to the buyer. After collecting all bids, the buyer selects a subset of  sellers $S$ as winners and determines the payment $p_i$ to each winner $s_i$. 

The proposed auction mechanism, MSensing, selects winners $S$ in a greedy manner. Starting with $S = \emptyset$, it iteratively chooses the seller that brings the largest non-negative net marginal profit.
Each winner $s_i \in S$ is paid the critical value $p_i$ of $s_i$, that is, seller $s_i$ would not win the auction if $s_i$ bids higher than $p_i$. 
Specifically,
MSensing runs the winner selection algorithm over users $S' = U \setminus \{s_i\}$. The payment $p_i$ is the largest price $s_i$ can bid, such that $s_i$ can replace a user in $S'$. Please note that $p_i \geq b_i$, this is because due to incomplete cost information, the buyer provides extra compensations to sellers on top of their bids to motivate them to reveal actual costs.
MSensing satisfies Myerson's characterization of truthful auction mechanisms~\citep{myerson1981optimal}.

The follow-up work by~\citet{jin2015quality} considers the situation where a data buyer has a data quality requirement $Q_j$ for each sensing task $t_j$. The authors propose a 
Vickrey-Clarke-Groves mechanism~\citep{ausubel2006lovely}
like truthful reverse combinatorial auction. They assume that the data quality $q_i$ of each seller $s_i$ is public and $q_i$ is the same for all sensing tasks. 
The authors first consider the scenario where each seller only bids for one bundle of sensing tasks $\Gamma_i$. The auction winners $S$ must satisfy the quality requirement for each task $t_j$, that is, $\sum_{s_i \in S, \text{ if } t_j \in \Gamma_i } q_i \geq Q_j$.
The objective of the auction is to maximize the total utility of the buyer and the sellers. 
The authors prove that winner determination under the setting is NP-hard and propose a greedy winner selection algorithm with a guaranteed approximation ratio to the optimal total utility. Each winner is paid by the winner's critical payment. 
The authors further study the total utility maximization problem in a more general scenario, where each seller can bid for multiple bundles of tasks. They propose an iterative descending algorithm that achieves close-to-optimal total utility. However, the auction is not truthful.

\citet{koutsopoulos2013optimal} considers a similar setting as~\citet{jin2015quality} do, but assumes that a data buyer has only one sensing task. The author proposes a truthful reverse auction that minimizes the expected cost of the buyer while guaranteeing the data quality requirement. The author assumes that the data buyer has prior knowledge about the distribution of each seller $s_i$'s unit participation cost $c_i$.
The units of participation $x_i$ of $s_i$ is a positive real value indicating how much data is purchased from $s_i$.
Given the sellers' bids, the data buyer determines the auction winners and their participation units by solving a linear programming model, which minimizes the total expected payment under the data quality constraint. Critical payments are made to the selected winners. All sellers bidding truthfully forms a Bayesian Nash equilibrium~\citep{DBLP:series/synthesis/2008Leyton}.

\subsection{Pricing Data Queries}
\label{sec:price_query}

Query-based pricing models tailor the purchase of data to users' needs. Customers can purchase their interested parts of a data set through data queries, and are charged according to their issued queries. While such a marketplace mechanism provides greater flexibility to buyers, a less carefully designed pricing model may open the loophole for arbitrage, which allows buyers to obtain a query result in a cost less than the advertised prices. 

Given a database $D$ and a multi-set of query bundles $\mathbf{S} = \{\mathbf{Q}_1, \ldots, \mathbf{Q}_m\}$, 
a query bundle $\mathbf{Q}$ is determined by $\mathbf{S}$, if the answer to $\mathbf{Q}$ can be computed only from the answers to the query bundles in $\mathbf{S}$. A pricing function is arbitrage-free if the advertised price $\pi (\mathbf{Q}) \leq \sum_{i=1}^{m} \pi (\mathbf{Q_i})$, that is, the answer to a query bundle $\mathbf{Q}$ cannot be obtained more cheaply from an alternative set of query bundles.

The first formal framework for arbitrage-free query-based data pricing was introduced by \citet{DBLP:conf/pods/KoutrisUBHS12}. 
The major idea is that a data seller can first specify the prices of a few views $\mathbf{V}$ over a database, and then the price of a query bundle $\mathbf{Q}$ is decided algorithmically. Theoretically, the authors show that if there are no arbitrage situations among the views in $\mathbf{V}$, there exists a unique arbitrage-free and discount-free pricing function $\pi(\mathbf{Q})$. Specifically, $\pi(\mathbf{Q})$ is the total price of the cheapest subset of $\mathbf{V}$ that determines $\mathbf{Q}$, which is found by query determinacy~\citep{DBLP:conf/icdt/NashSV07}. They also show the complexity of evaluating the price functions. Unfortunately, the pricing model is NP-hard for a large class of practical queries. They develop polynomial time algorithms for specific classes of conjunctive queries, chain queries, and cyclic queries.


Subsequently, \citet{DBLP:conf/sigmod/KoutrisUBHS13} develop a prototype pricing system, QueryMarket, based on the idea~\citep{DBLP:conf/pods/KoutrisUBHS12}. They formulate the pricing model as an integer linear program (ILP) with the objective to minimize the total cost of purchased views $\mathbf{V}_{p}$. 
The purchased views $\mathbf{V}_{p}$ must satisfy the following requirements. For a tuple $t$ in the query answer $\mathbf{Q}(D)$, there must exist a subset of views in $\mathbf{V}_{p}$ that can produce $t$ and for each relation $R$ in $\mathbf{Q}$, at lease one view on $R$ should be purchased. For a tuple $t$ not in $\mathbf{Q}(D)$, there must exist a subset of views in $\mathbf{V}_{p}$ that can indicate $t \notin \mathbf{Q}(D)$.
Although the pricing problem in the setting is in general NP-hard, QueryMarket shows that a large class of queries can be priced in practice, albeit for small data sets. To handle the case that a query $\mathbf{Q}$ may require databases from multiple sellers, they introduce a revenue sharing policy among sellers. Specifically, each seller gets a share of the query price $\pi(\mathbf{Q})$, which is proportional to the maximum revenue that the seller can get among all minimum-cost solutions to the ILP.


The problem of designing arbitrage-free pricing models for linear aggregation queries is studied by~\citet{DBLP:conf/icdt/LiLMS13}. Given a data set of $n$ real values $\mathbf{x} = \langle x_1, \ldots, x_n \rangle$, a linear query over $\mathbf{x}$ is a real-valued vector $\mathbf{q}=\langle w_1, \ldots, w_n \rangle$, and the answer is $\mathbf{q}(\mathbf{x})=\sum_{i=1} w_i x_i$.   The authors propose a marketplace, where a data buyer can purchase a single linear query $\mathbf{q}$ with a variance constraint $v$ defined by the buyer. The query $\mathbf{Q}=(\mathbf{q}, v)$ is answered by an unbiased estimator of $\mathbf{q}(\mathbf{x})$ with a variance smaller than or equal to $v$.
The authors first develop a proposition that the pricing function $\pi$ cannot decrease faster than $\frac{1}{v}$, that is, $\pi(\mathbf{q}, v) = \Omega(\frac{1}{v})$. Then, they propose a family of arbitrage-free pricing functions, $\pi(\mathbf{q}, v) = \frac{f^2(\mathbf{q})}{v}$, where the function $f(\cdot)$ is semi-norm. 
Last, they provide a general framework to synthesize new arbitrage-free pricing functions from the existing ones. For any arbitrage-free pricing functions $\pi_1, \ldots, \pi_k$, the pricing function $\pi(\mathbf{Q})=f(\pi_1(\mathbf{Q}), \ldots, \pi_k(\mathbf{Q}))$ is also arbitrage-free if $f(\cdot)$ is a subadditive and nondecreasing function. 
A comprehensive list of celebrated arbitrage-free pricing functions are listed by~\citet{DBLP:conf/kdd/NiuZWTGC18}.
In addition to synthesized pricing functions,~\citet{DBLP:conf/icdt/LiLMS13} also study a similar view-based pricing framework as~\citet{DBLP:conf/pods/KoutrisUBHS12} do.
By adapting the theoretical results in~\citep{DBLP:conf/pods/KoutrisUBHS12}, the authors show that the view-based pricing model for linear aggregation queries is NP-hard.


\citet{DBLP:journals/pvldb/LinK14} study arbitrage-free pricing for general data queries. They propose three pricing schemes, namely instance-independent pricing, up-front dependent pricing, and delayed pricing. 
The authors further summarize five forms of arbitrages, namely price-based arbitrage, separate account arbitrage, post-processing arbitrage, serendipitous arbitrage, and almost-certain arbitrage. The authors point out that the model by \citet{DBLP:conf/pods/KoutrisUBHS12} has pricing-based arbitrage, that is, the computed prices may leak information about $D$. Theoretically,
they propose an instance-independent pricing function and a delayed pricing function that are arbitrage-free across all forms. The major idea is to tackle the pricing problem from a probabilistic view. Queries that are more likely to reveal the true database instance are priced higher.


In the same vein, \citet{DBLP:conf/icdt/DeepK17} characterize the structure of pricing functions with respect to information arbitrage and bundle arbitrage, where information arbitrage covers both post-processing arbitrage and serendipitous arbitrage defined by~\citet{DBLP:journals/pvldb/LinK14}. For both instance-independent pricing and answer-dependent pricing of a query, an arbitrage-free pricing function should be monotone and subadditive with respect to the amount of information revealed by asking the query. Several examples of arbitrage-free pricing functions are presented, including the weighted coverage function and the Shannon entropy function.


\citet{DBLP:conf/sigmod/DeepK17} later implement the theoretical framework~\citep{DBLP:conf/icdt/DeepK17} into a real time pricing system, QIRANA, which computes the price of a query bundle $\mathbf{Q}$ from the view of uncertainty reduction. They assume that a buyer is facing a set of all possible database instances $S$ with the same schema as the true database instance $D$. After receiving the query answer $E=\mathbf{Q}(D)$, the buyer can rule out some database instances $D_i \in S$ that cannot be $D$ by checking whether $\mathbf{Q}(D_i) = E$. A query bundle that eliminates more database instances is priced higher, as it reveals more information about $D$. The authors propose an arbitrage-free answer-dependent pricing function, which assigns a weight $w_{i}$ to each database $D_i \in S$, and computes the price of a query bundle by 
\begin{equation}
    \pi(\mathbf{Q})=\sum_{i \in \{i | D_i \in S\}, \mathbf{Q}(D) \neq \mathbf{Q}(D_i)} w_i.
    \label{eq:QIRANA}
\end{equation}
By default, the same weight $w_i = \frac{P}{|S|}$ is assigned to each possible database instance $D_i$, where $P$ is a parameter set by the data owner. The data owner can also provide QIRANA with some example query bundles and their corresponding prices. Then, QIRANA will automatically learn instance weights $w_i$ from the given examples by solving an entropy maximization problem.
Choosing $S$ to be the complete set of possible database instances leads to a $\#P$-hard problem. To make the pricing function tractable, QIRANA uses a random sample of database instances as $S$.


\citet{DBLP:journals/pvldb/ChawlaDKT19} extend the pricing function in Equation~\ref{eq:QIRANA} to maximize seller revenue. They consider the setting that the supply is unlimited and the buyers are single-minded, that is, a buyer only wants to buy a single query bundle $\mathbf{Q}$. A buyer will purchase $\mathbf{Q}$ if the advertised price $\pi(\mathbf{Q})$ is smaller than or equal to the buyer's
valuation $v_{\mathbf{Q}}$. The authors take a training data set consisting of some query bundles and their customer valuations. Three pricing schemes are investigated. The major idea of the pricing schemes is that, according to Equation~\ref{eq:QIRANA}, a query can be priced as a bundle of items (database instances).
Uniform bundle pricing sets the same price for all query bundles. Item pricing sets the price of a query bundle using Equation~\ref{eq:QIRANA}, where the weights $w_{i}$ are learned from the training data. XOS pricing learns $k$ weights $w_i^1, \ldots, w_i^k$ for each item $D_i$ and sets the price of $\mathbf{Q}$ as $\pi(\mathbf{Q})=\text{max}_{j=1}^{k}\sum_{i \in \{i|D_i \in S\}, \mathbf{Q}(D) \neq \mathbf{Q}(D_i)} w_i^j$. Theoretically, the approximation rate of each pricing scheme to the optimal revenue is studied. Although XOS pricing scheme enjoys the best approximation rate, the authors show that item pricing usually achieves larger revenue in practice.


\citet{miao2020towards} study the problem of pricing selection-projection-natural join queries over incomplete databases. An arbitrage-free pricing function is proposed based on the idea of data provenance, which describes the origins of a piece of data and its processing history~\citep{DBLP:conf/sigmod/BunemanT07, DBLP:conf/dexa/TangWBBV13}. Let $t$ be a tuple in a query answer $\mathbf{Q}(D)$.
The lineage $L(t, D)$ of  $t$ is defined as the set of tuples in the database  $D$ that contribute to $t$. The authors assume that each tuple $t$ has a base price $p(t)$.
The price of $\mathbf{Q}$ is set to the weighted aggregation of the costs of all tuples in $M(\mathbf{Q}, D)=\cup_{t \in \mathbf{Q}(D)}L(t, D)$. Specifically, $\pi^{UCA}(\mathbf{Q}) = \sum_{i \in \{i|t_i \in M(\mathbf{Q}, D)\}} \mu_i  p(t_i)$, where $\mu_i$ is the percentage of attributes of $t_i$ that are not missing. The authors also propose an answer quality aware pricing function, $\pi^{QUCA}(\mathbf{Q})=\frac{\Delta}{n} \pi^{UCA}(\mathbf{Q})  \kappa(\mathbf{Q}, D)$, where $\kappa(\mathbf{Q}, D)$ is the answer quality and $\Delta$ is a constant. However, $\pi^{QUCA}$ is not arbitrage-free.


Purchasing data is usually not a one-shot deal. A customer may purchase multiple queries from the same data seller. A history-aware pricing function will not charge the customer twice for already purchased information. QueryMarket~\citep{DBLP:conf/sigmod/KoutrisUBHS13} tracks the purchased views of a customer and avoids charging those views when pricing future queries of the customer. Both \citep{DBLP:conf/sigmod/DeepK17} and~\citep{miao2020towards} support history-aware pricing in the same vein as~\citep{DBLP:conf/sigmod/KoutrisUBHS13}. One drawback of these history-based approaches is that the seller must provide reliable storage to keep users' query history~\citep{DBLP:journals/pvldb/UpadhyayaBS16}. 


\citet{DBLP:journals/pvldb/UpadhyayaBS16} propose an optimal history-aware pricing function, that is, a buyer is only charged once for purchased data. 
The key idea is to allow buyers to ask for refunds of already purchased data. In their setting, a query is priced according to its output size. 
The seller computes an identifier (coupon) for each tuple in the query answer $\mathbf{Q}(D)$. Both $\mathbf{Q}(D)$ and the corresponding coupons are sent to the buyer.
If the buyer receives the same tuple $t$ from two  queries, the buyer can ask for a refund of $t$ by presenting the two coupons associated with $t$ in the two corresponding queries. 
To prevent buyers from borrowing coupons from others and receiving unconscionable refunds, each coupon is uniquely associated with a buyer.
By tracking coupon status, the data seller guarantees that each coupon will be used only once.
However, the pricing function has no arbitrage-free guarantee~\citep{DBLP:conf/sigmod/DeepK17}.

\subsection{Privacy Compensation}
Machine learning models in many areas, like recommendation systems~\citep{DBLP:conf/wine/DandekarFI12} and personalized medical treatments~\citep{DBLP:conf/aaai/MaZWRWTMGG20}, require a large amount of personal data. However, trading and sharing personal data may leak the privacy of data providers. Therefore, how to measure and properly compensate data providers for their privacy loss is an important concern in designing marketplaces of personal data.

Differential privacy~\citep{DBLP:conf/tamc/Dwork08}  is a mathematical framework rigorously providing privacy protection and plays an essential role in personal data pricing.
Following the principle of differential privacy, random noises are injected into a data set, such that data buyers can learn useful information about the whole data set but cannot learn specifics accurately about an individual. 
The magnitude of random noise impacts data providers' privacy loss and the data price. A data set with less injected random noise may leak more privacy and is priced higher. Pricing models of personal data routinely adopt cost-plus pricing strategy, where sellers first compensate data providers for their privacy loss, and then scale up the total privacy compensation to determine the price for data buyers~\citep{DBLP:conf/kdd/NiuZWTGC18}.


\citet{DBLP:conf/sigecom/GhoshR11} initiate the study of pricing privacy by auction.
They propose a truthful marketplace to sell single counting queries on binary data.
In their settings, a data seller has a data set consisting of personal data $d_i \in \{0, 1\}$ of individual $i$. 
The data seller sells an estimator $\widehat{s}$ of the sum $s=\sum_{i}d_i$ and compensates data providers for their privacy loss. 
Under the framework of differential privacy, the authors treat privacy as a commodity to be traded. In particular, if a provider's data is used in an $\epsilon-$differentially private manner, $\epsilon$ privacy units should be purchased from the provider. Thus, the privacy compensation problem can be transformed into variants of multi-unit reverse auction.
The authors assume that each data provider $i$ has a privacy cost function 
\begin{equation}
c_i(\epsilon) = v_i*\epsilon,    
\label{eq:ghosh_privacy_cost}
\end{equation}
representing the cost for using the data in an $\epsilon$-differentially private manner, where $v_i$ is the unit privacy cost of $i$.
In an auction, data providers are asked to submit their asking prices $b_i$ for the use of their data. 
\citet{DBLP:conf/sigecom/GhoshR11} consider two situations. In the first situation, a buyer has an accuracy requirement on $\widehat{s}$, that is, $\text{Pr}[|\widehat{s} - s| \geq k] \leq \frac{1}{3}$. The authors establish an observation that they only need to purchase data from $m$ individuals and use them in an $\epsilon$-dfferential privacy manner, where $m$ and $\epsilon$ only depend on the accuracy goal. 
It's shown that the classic Vickrey-Clarke-Groves auction minimizes the buyer's payment and guarantees the accuracy goal.
The major idea is to select $m$ individuals with the cheapest bids and provide each winner with a uniform compensation $\epsilon \cdot b$, where $b$ is the $(m+1)$-th smallest bid. In the second situation, a buyer has a budget constraint and wants to maximize the accuracy of $\hat{s}$. The authors propose a greedy-based approximation algorithm to solve the problem. 


The value of personal data and privacy valuation may be correlated. For example, a patient may assign a higher price to the patient's medical report than the healthy people ask for.
\citet{DBLP:conf/sigecom/GhoshR11} show a negative result that in the situations having such correlations, no individually rational direct mechanism can protect privacy.


In a follow-up study, \citet{DBLP:conf/wine/DandekarFI12} consider the scenario of selling linear aggregate queries $\mathbf{q}=\langle w_1, \ldots, w_n \rangle$ over real-valued personal data $D=\langle d_1, \ldots, d_n \rangle$.
They assume data providers have the same privacy cost function as Equation~\ref{eq:ghosh_privacy_cost}, and propose a truthful reverse auction mechanism to maximize the accuracy of estimators for budget-constraint buyers.
The error of an estimator $\widehat{s}$ of the true answer $s=\sum_{i} w_i  d_i$ is its squared error $(\widehat{s} - s)^2$. 
It is shown that an $\widehat{s}$ computed from more providers with large corresponding weights in $\mathbf{q}$ is more accurate.
Therefore, the problem is transformed into a knapsack reverse auction~\citep{DBLP:conf/focs/Singer10} that maximizes the total weights of the selected providers under budget constraints. Specifically, the authors treat the budget as the capacity of the knapsack, the privacy cost of a data entry $d_i$ as its weight in the knapsack, and $w_i$ as the value of $d_i$. A greedy-based algorithm with an approximation ratio of $5$
is proposed to solve the problem. 


The aforementioned studies~\citep{DBLP:conf/sigecom/GhoshR11,DBLP:conf/wine/DandekarFI12} assume that data buyers can purchase an arbitrary amount of privacy from each data provider. However, a conservative individual may not want to sell the individual's data if the privacy loss is too large. \citet{DBLP:conf/sigir/NgetCY17} study the same problem as~\citet{DBLP:conf/wine/DandekarFI12} do in a more realistic situation, that is, an individual $i$ can refuse to participate in an estimator if the privacy loss of $i$ is larger than a threshold $\epsilon_i$. They assume that the privacy cost function of each data provider is public and propose a heuristic method to determine query price. The model first randomly samples a subset of data providers. Then, it uses the data from each sampled individual $i$ in an $\epsilon_i$-differentially private manner, and computes the compensations correspondingly. If the total compensation is larger than the budget,
the model decreases the differential privacy levels of the high cost providers, such that the budget goal is met. Last, they generate the perturbed query answers by personalized differential privacy~\citep{DBLP:conf/icde/JorgensenYC15}, which guarantees the differential privacy for each selected individual $i$. They repeat the above steps several times and return the perturbed answer with the smallest squared error.

Later, \citet{DBLP:conf/uai/ZhangBL20} propose a truthful personal data marketplace, where each data provider $i$ can specify the personal maximum tolerable privacy loss $\epsilon_i$.
They first show that the accuracy of query answers is proportional to the total amount of purchased privacy. 
Under the assumption that the distributions of privacy costs of all individuals are public, they design a variant of Bayesian optimal knapsack procurement~\citep{DBLP:journals/eor/EnsthalerG14}, which maximizes the expected total purchased privacy under the constraint of a data buyer's expected budget. The authors solve the problem by adopting the algorithm in~\citep{DBLP:journals/eor/EnsthalerG14}. The noisy query answer is generated using personalized differential privacy~\citep{DBLP:conf/icde/JorgensenYC15}, which guarantees $\epsilon_i$-differential privacy for each selected individual $i$.

The models proposed by \citet{DBLP:conf/wine/DandekarFI12, DBLP:conf/sigecom/GhoshR11} may be attacked by arbitrage. 
\citet{DBLP:conf/icdt/LiLMS13} consider the situation where a data buyer has a variance constraint $v$ on the purchased noisy query answers. 
They assume that the privacy costs of individuals are public, and propose a theoretical framework for assigning arbitrage-free prices to linear aggregate queries $\mathbf{q}$. 
A perturbed answer is generated from the true answer by adding Laplace noise with the expectation $0$ and variance $\sqrt{\frac{v}{2}}$. Measured by differential privacy, the privacy loss of an individual $i$ is upper-bounded by $\epsilon = \frac{w}{\sqrt{\frac{v}{2}}}$
if the individual is involved in the query, and $0$ otherwise, where $w$ is the largest absolute weight in $\mathbf{q}$.
Several privacy compensation functions are proposed, such as $p_i(\epsilon) = c_i \epsilon$, where $c_i$ is the unit privacy cost of individual $i$. The price of a query is the sum of the privacy compensations, which is proved to be arbitrage-free.

\citet{DBLP:conf/icdt/LiLMS13} only compensate individuals involved in queries. However, as two individuals' data may be correlated, the privacy of a not-involved individual may be leaked due to the revelation of the other individual's data. To fairly compensate individuals for their privacy, \citet{DBLP:conf/kdd/NiuZWTGC18} extend the model by~\citet{DBLP:conf/icdt/LiLMS13}, and propose a pricing model that is arbitrage-free and dependency fair. 
Dependent fairness requires that a data provider should receive a privacy compensation as long as some data of other providers that is correlated to the data of this provider is involved in a query.
Employing dependent differential privacy~\citep{DBLP:conf/ndss/LiuMC16}, the privacy loss of a data provider $i$ caused by a query is upper-bounded by $\epsilon_i = \frac{ds_i}{\sqrt{\frac{v}{2}}}$, where $ds_i$ is the dependent sensitivity of the query at provider $i$'s data. The authors propose a bottom-up mechanism and a top-down mechanism to determine privacy compensations and query prices. The bottom-up mechanism  computes compensations in the same way as~\citet{DBLP:conf/icdt/LiLMS13} do and determines query prices as a multiple of the total compensations. The top-down mechanism first determines the query price using a user-defined arbitrage-free pricing function
and spares some fraction of a buyer's payment for privacy compensation. Each data provider receives a division of the compensation proportional to the provider's privacy loss. 


All of the privacy compensation methods discussed above assume a trustworthy platform/agent to trade data providers' privacy with data buyers. Data providers, however, cannot control the usage of their own data. 

In this concern,~\citet{DBLP:conf/infocom/JinXLG19} develop a truthful crowdsensing marketplace, where data owners can determine how much privacy to disclose.
In their marketplace, obfuscated geo-locations of data owners are traded by auctions. Data owners first inject random noise to their data based on their own privacy preferences. Then, each data owner bids with the cost as well as the mean and variance of the injected random noise.
The buyer determines auction winners to maximize data accuracy with respect to the buyer's budget constraint. The authors show that the optimization problem is NP-hard and develop a greedy heuristic solution. The major idea is to iteratively select data owners that bring the largest marginal utility contributions until the budget is used up. 


\bigskip

In this section, we review representative pricing models of raw data sets in four types of scenarios, where different desiderata are considered.
A limitation of the discussed pricing models is that data sets are priced without considering their down-stream applications. \citet{10.14778/3407790.3407800} argue that the value of a data set to customers is usually task dependent and cannot be evaluated by the intrinsic properties of the data set alone. 
As the pricing models of raw data sets are agnostic to the down-stream applications of raw data sets, these pricing models can be used in machine learning pipelines of building both supervised and unsupervised machine learning models.


\section{Pricing Data Labels}
\label{sec:price_label}

Crowdsourcing is a popular method for collecting large-scale labeled training data for machine learning tasks~\citep{shah2016double}. Unfortunately, crowdsourced data often suffers from quality issues. This is mainly due to the existence of lazy and spamming workers, who submit low quality labels. Those workers can be discouraged from participating in the tasks by rewarding them with a performance-based payment~\citep{radanovic2016incentives}. However, due to a lack of ground-truth verification of the collected labels, how to evaluate label quality and price the labels correspondingly is a challenging task.
In this section, we review two types of label pricing models, which are designed to motivate workers to exert efforts and submit accurate data labels.

\subsection{Gold Task-based Pricing Models}

A gold task is one for which the answer is known to the data buyer a priori.  Gold tasks can be uniformly mixed at random within the tasks for workers to evaluate workers' performance, which determines the payments to workers. Since workers cannot distinguish gold tasks from others, this strategy can motivate workers to provide accurate labels.


\citet{shah2016double} consider a crowdsourcing setup where workers perform binary labeling tasks. The authors propose a multiplicative pricing model  using gold tasks. The model allows a worker to skip an assigned task if the worker is not confident about the answer. The total payment to a worker $u$ is computed based on $u$'s performance on the answered tasks. 
The workers are selfish and want to maximize their individual expected payments. The authors assume that each worker has a private belief $\text{Pr}(y_{t}=l)$ about how likely the true label $y_{t}$ of a task $t$ is $l$. The pricing model is designed to incentivize workers to only report high-confidence labels with beliefs greater than a threshold $p$. The total reward starts at $\beta$.  For each correct answer in the gold tasks, the reward will be multiplied by $\frac{1}{p}$. However, if any of these gold tasks are answered incorrectly, the reward will drop to zero, that is, 
\begin{equation}
    \pi(u) = \beta \cdot \frac{1}{p^{c}} \cdot \mathbf{1}(r=0),
    \label{eq:gold_task_double}
\end{equation}
where $\mathbf{1}(\cdot)$ is an indicator function, and $c$ and $r$ are the number of correct and wrong answers, respectively. This pricing model motivates workers to only answer tasks that they are sufficiently confident about. 
The pricing model is incentive compatible, that is, a worker receives the maximum expected payment if and only if the worker exerts efforts to report accurate labels. The pricing model also satisfies the ``no-free-lunch'' axiom, that is, workers who only provide wrong answers will receive no payments. 
In their setting, the proposed method is the unique incentive compatible model that satisfies the ``no-free-lunch'' axiom. 


\citet{DBLP:conf/icml/ShahZP15} further generalize the model~\citep{shah2016double} to multi-label tasks.
For each task, a worker can submit multiple answers $\widehat{Y}$ that the worker believes is most likely to be correct. 
This multi-selection system provides workers more flexibility to express their beliefs, which can use the expertise of workers with partial knowledge more effectively than single-selection systems.
The authors assume that the workers' beliefs for any label being the true label for a task lie in the set $\{0\} \cup (p, 1]$, where $p$ is fixed and known.
The authors want to encourage workers to only report the set of labels with positive beliefs.
The reward of a worker for a gold task is $(1-p)^{(|\widehat{Y}|-1)}$ if one of the worker's answers is correct and $0$ if otherwise. 
The total payment to a worker is determined by the product of the worker's rewards on all gold tasks. 


In a later study, \citet{DBLP:conf/icml/ShahZ16} propose a two-stage multiplicative pricing model to motivate workers to self-correct their answers. In the first stage, a worker answers the assigned tasks. 
In the second stage, if the worker's answer to a task $t$ does not agree with the answer from the peer workers, the worker has an opportunity to change the answer. The worker $u$ receives a high reward for a gold task $t$ if the initial answer to the task is correct, a low reward if the updated answer is correct, and $0$ reward if the final answer is wrong. The total payment is determined by the product of the worker's rewards from gold tasks. 
Theoretically, the authors prove that the proposed method is the unique incentive compatible model that satisfies the no-free-lunch axiom.
Empirically, they show in a simulation that the self-correction setting can significantly improve the data quality compared to the standard single-stage settings. 


To reduce the variance in payoffs, the aforementioned methods~\citep{DBLP:conf/icml/ShahZ16, DBLP:conf/icml/ShahZP15, shah2016double} require each worker to solve a sufficient number of gold tasks. This leads to a waste of procurement budget, as the answers to the gold tasks are already known.

\citet{de2016incentives} address the limitation by combining the ideas from peer prediction and gold tasks. They arrange the workers in a hierarchy, where every worker shares one common task with each of its children.
A few gold tasks are used to incentivize high efforts from the workers at the top level of the hierarchy. 
Assuming these workers exert sufficient efforts to provide high quality answers, their answers can be used as pseudo gold tasks for workers in the second layer, who can in turn provide pseudo gold tasks for the next level, and so forth. 
A worker will be punished if the worker does not agree with the parent on the task shared between them. As the workers at the top level are evaluated by the true gold tasks, they are evaluated more accurately than the other workers, which is not fair to workers at lower layers.


The follow-up work by \citet{DBLP:conf/aaai/GoelF19} considers fair payment among workers, that is, the expected reward of a worker is directly proportional to the accuracy of the worker's answers and independent of the strategy and proficiency of the worker's random peers. 
The key idea is to estimate the proficiency of workers, which is the probability that a worker can solve the tasks correctly. 
\citet{DBLP:conf/aaai/GoelF19} start by estimating the proficiency of a small group of workers with gold tasks. 
Then, the answers by the small group of workers to non-gold tasks are used as contributed gold tasks, where the workers' proficiencies are used as the trustworthy degree of those tasks. The contributed gold tasks are used to estimate the proficiency of more workers. Finally, the payoff of each worker is proportional to the worker's estimated proficiency, such that workers with good proficiency receive high payments. The model guarantees that exerting high efforts to provide accurate labels is a dominant strategy for each worker.

\subsection{Peer Prediction-based Pricing Models}

Peer prediction-based pricing model can incentivize efforts and accurate data labels without access to gold tasks.
Those models take advantage of the stochastic correlation of answers to the same tasks, and set up a game among workers, called a mechanism in game theory. 
The game is designed such that workers who exert effort in solving the tasks can achieve high expected rewards, whereas spammers providing random answers on average receive no payments. A pricing model is incentive compatible if it admits exerting high efforts and truthful reporting as an equilibrium.


\citet{DBLP:conf/www/DasguptaG13} initiate the study of effort elicitation and propose the DG model to price binary labels. 
A data buyer assigns a set of data labeling tasks to a group of workers, such that each task is labeled by multiple workers and each worker labels multiple tasks. They assume that a worker $u_i$ either invests no effort and thus provides a random label, or invests full effort with a cost $c_i$ and provides a true label with probability $p_i$. Here, $p_i$ is called the proficiency of $u_i$.  The workers are self-interested, who want to maximize their payoffs.

The DG model pays a worker $u_i$ on an assigned task $t$ based on how surprisingly $u_i$'s report is consistent with that of the peer worker $u_p$. Denote by $\widehat{y}$ and $\widehat{y}_p$ the answers from $u_i$ and $u_p$ to a task, respectively.
The model pays $u_i$ with a constant reward subtracting the probability $\text{Pr}(u_i, u_p)$ that $u_i$ and $u_p$ have the same answer to a random task, that is,
\begin{equation}
    \pi(u_i, t) = \beta \cdot (\mathbf{1}(\widehat{y} = \widehat{y}_{p}) - \text{Pr}(u_i, u_p)),
    \label{eq:peer_predict_binary}
\end{equation}
where  $\beta$ is a non-negative payment scaling parameter that is chosen to cover workers' effort costs, and $\text{Pr}(u_i, u_p)$ is approximated from the submitted labels. The total payment to a worker $u_i$ is the sum of $u_i$'s payment for each task.

The pricing model incentivizes efforts, as the expected payment for spammers who do not solve their tasks and report random/constant labels is exactly zero. Under the assumption that the proficiency of all workers are better than random guess,
it is shown that the DG model is incentive compatible.
Even though the pricing model also has non-informive equilibria, such as all workers reporting the same label, those equilibria are less profitable to the workers, and thus are not attractive to the workers.


In a multi-label situation, two labels $l_1$ and $l_2$ may be positively correlated. \citet{DBLP:conf/sigecom/ShnayderAFP16} show that under the DG model~\citep{DBLP:conf/www/DasguptaG13}, workers can achieve more profits by misreporting $l_1$ by $l_2$. The correlated agreement (CA) mechanism~\citep{DBLP:conf/sigecom/ShnayderAFP16} extends the DG model to multi-label tasks. In the CA mechanism, knowledge about label correlation is required.
A label correlation matrix $\Delta$ is learned from workers' submissions, where an element $\Delta_{i, j}= \text{Pr}(l_i, l_j) - \text{Pr}(l_i)\text{Pr}(l_j)$ is the correlation degree between labels $l_i$ and $l_j$. Denote by $S(\cdot)$ the sign function of $\Delta$, that is, $S(l_i, l_j)=1$ if $\Delta_{i, j} > 0$, and $0$ otherwise.
A worker $u$ will be rewarded for a task $t$ if $u$'s report is positively correlated with that of peer $u_p$. To penalize the case where all workers blindly report the same label, a worker $u$ will be penalized if $u$ is likely to be consistent with worker $u_p$ on random tasks. In particular, the payment to worker $u$ for reporting $\hat{y}$ is
\begin{equation*}
    \pi(u, t) = \beta \cdot (S(\widehat{y}, \widehat{y}_p) - S(\widehat{y}_{a}, \widehat{y}_b)),
\end{equation*}
where $\hat{y}_p$ is the answer to task $t$ by worker $u_p$, $\widehat{y}_{a}$ is the answer to a random task by worker $u$, and $\widehat{y}_b$ is the answer to another random task by worker $u_p$. When the number of tasks is large, such that label correlations $\Delta$ can be accurately learned, the CA mechanism is incentive compatible with the highest payment. However, the mechanism fails if two labels $l_1$ and $l_2$ are not distinguishable with respect to $S(\cdot)$, that is, $\forall l_i \in Y$, $S(l_1, l_i) = S(l_2, l_i)$. In this situation, workers may misreport $l_1$ by $l_2$ and still receive the same payoffs.


\citet{radanovic2016incentives} provide complementary theoretical results on pricing multi-label tasks. They assume that the labels only have limited correlations, that is, $\text{Pr}(o_p = l_2 |o = l_1) < \text{Pr}(o_p = l_2 |o=  l_2)$, where $o$ and $o_p$ are the observed labels of worker $u$ and worker $u_p$, respectively. The mechanism pays the report $\widehat{y}$ by worker $u$  on a task $t$ by
\begin{equation*}
    \pi(u, t) = \frac{\mathbf{1}(\widehat{y}=\widehat{y}_p)}{\text{R}(\widehat{y})} - 1,
    \label{eq:peer_ptsc}
\end{equation*}
where $\text{R}(\widehat{y})$ is the empirical frequency of $\widehat{y}$, which is computed from all submissions. It is shown that exerting high efforts and truthful reporting is strictly more profitable than any other equilibria. However, their assumptions on label correlations may not hold in some applications~\citep{DBLP:conf/sigecom/ShnayderAFP16}.


The aforementioned methods~\citep{DBLP:conf/www/DasguptaG13, DBLP:conf/sigecom/ShnayderAFP16, radanovic2016incentives} require that each task must be completed by at least two workers, which leads to duplicate answers, and thus does not use the crowd efficiently. For a setting with binary labels, \citet{liu2017machine} propose to learn a classifier $\mathcal{M}$ from workers' reports, and use the classifier's predictions $\mathcal{M}(t)$ as peer reports. Since workers' submitted labels are noisy, the classifier is trained by the techniques of learning with noisy labels~\citep{DBLP:conf/nips/NatarajanDRT13}. Specifically, they first estimate the error rates of submitted labels. Then, the classifier is optimized by an error rate calibrated loss function $\varphi(\cdot)$ proposed by~\citet{DBLP:conf/nips/NatarajanDRT13}. A report $\widehat{y}$ to a task $t$ is priced based on $-\varphi(\mathcal{M}(t), \widehat{y})$, such that labels with large loss are priced lower. 
Under the assumption that $\mathcal{M}$ is better than random guess, exerting efforts to find the truth labels is the highest-paying equilibrium.


\citet{DBLP:conf/aaai/LiuC17} study the problem of sequential label collection, where labeling tasks are published in multiple rounds. 
In their settings, an accurately labeled task has a fixed reward to a data buyer, whereas a mistakenly labeled task has no value to a data buyer.
They propose an incentive compatible pricing model that maximizes the expected utility for a data buyer, which is the difference between the total rewards and the total payment.

They develop a multi-armed bandit algorithm to extend the DG model~\citep{DBLP:conf/www/DasguptaG13}, which dynamically adjusts the parameter $\beta$ in Equation~\ref{eq:peer_predict_binary}. 
A larger $\beta$ encourages more accurate labels but costs more money.
As the bandit algorithm requires a static environment, this method may fail to learn the optimal $\beta$ if adversarial workers adjust their strategies according to their interactions with the mechanism~\citep{DBLP:conf/nips/HuLZLL18, DBLP:conf/nips/GurZB14}. \citet{DBLP:conf/nips/HuLZLL18} solve the problem by reinforcement learning, which is more robust to strategic behaviors of workers.



In practice, peer prediction-based models need to adjust payments to avoid negative payments. The adjustment may lead to an issue that spammers may receive positive and high rewards.
\citet{radanovic2016learning} address the issue by proposing a reputation system PropeRBoost to adjust the payments. PropeRBoost publishes tasks to workers in multiple rounds, and computes a reputation score for each worker based on the worker's past submissions. In each round $r$, it first applies the DG model~\citep{DBLP:conf/www/DasguptaG13} to compute workers' payments, and then re-scales the payments by the reputations of the corresponding workers. It is shown that the average payment of a spammer converges to $0$ as $r$ approaches infinity.

\bigskip

In this section, we review gold task-based and peer prediction-based pricing models for data labels. The developed pricing models guarantee that exerting efforts to report accurate data labels is the most profitable strategy of all workers. A major concern of gold task-based methods is that these methods require a sufficient number of gold tasks to obtain good performance. In some scenarios, however, gold tasks are very expensive to obtain. For peer prediction-based methods, the existence of multiple equilibria is a major limitation, as workers may converge to an uninformative equilibrium, where workers do not exert full efforts~\citep{DBLP:conf/ijcai/ShnayderFP16}.

\section{Pricing in Collaborative Training of Machine Learning Models}
\label{sec:price_collaborative}

Collaborative machine learning is an appealing paradigm where multiple data owners collaboratively build high-quality machine learning models by contributing their data. 
As the data sets from different data owners may have different contributions to the trained machine learning models, data owners who contribute more valuable data should receive more rewards~\citep{sim2020collaborative}.
In this section, we review contribution evaluation and revenue allocation techniques in collaborative machine learning.

\subsection{Revenue Allocation by Shapley Value}
\label{sec:allocation_shapley}

Shapley fairness is widely adopted as the foundation of fair revenue allocation in collaborative machine learning. 
It guarantees that each participant receives a payment proportional to the participant's marginal contribution to the performance of the trained machine learning model. 
The challenge in adopting Shapley value lies in its exponential computational cost.


\citet{maleki2013bounding} tackle the efficiency issue of Shapley value by proposing a permutation sampling algorithm for bounded utility functions. By Equation~\ref{eq:shapley_2}, the Shapley value of a seller is the marginal utility contribution averaged over all possible subsets of sellers, which can be estimated by sample mean. Denote by $\widehat{\psi}(s)$ an $(\epsilon, \delta)$-approximator of a seller's Shapley value, that is, $\text{Pr}(|\widehat{\psi}(s) - \psi(s)| \leq \epsilon) \geq 1 - \delta$. To compute the estimators for all sellers, by Hoeffding's inequality~\citep{hoeffding1994probability}, we need $O(\frac{2r^2N}{\epsilon^2}\log\frac{2N}{\delta})$ samples and evaluate the utility function $O(N^2 \log N)$ times, where $N$ is the number of sellers and $r$ is the range of the utility function. 
Evaluating the utility function itself, such as computing testing accuracy, is computationally expensive, as it requires training a machine learning model. Therefore, the method is not scalable to a large number of sellers.


\citet{ghorbani2019data} extend the Monte-Carlo method by~\citet{maleki2013bounding} to price individual data point in supervised learning, and propose truncated-based and gradient-based approximation methods. Their truncated-based method reduces the number of utility evaluations by ignoring coalitions of large size. The authors argue that it is sufficient to estimate Shapley values up to the intrinsic noise in the prediction performance $\mathcal{U}$ on the test data set, which can be measured as the bootstrap variance of $\mathcal{U}$. In addition, the performance change by adding one more training data point $s$ to a large training data set $S$ is ignorably small. 
Therefore, if the utility of $S$ is close to the utility of the whole data set $D$, 
the marginal contribution of $s$ to $S$ can be regarded as $0$ in practice, and thus its computation can be truncated.
Their gradient-based method speeds up the evaluation of utility functions by reducing training time, where a model is trained with only one pass through the training data. They update the model by performing gradient descent on one data point $s$ at a time and the marginal contribution of $s$ is the change in the model performance. 
The two approximation methods introduce estimation bias into the approximated Shapley values, and have no guarantees on the approximation error.


\citet{jia2019towards} propose two approximation algorithms with provable error bounds for Shapley value that significantly reduce the number of utility evaluations. The first algorithm adopts the idea of group testing in feature selection~\citep{DBLP:conf/nips/ZhouPZNNRG14}. Denote by $\beta_i$ a Boolean random variable indicating whether a seller $s_i$ is in a random sample of sellers. A sampling distribution of $\beta_1, \ldots, \beta_N$ is designed such that the difference in Shapley values between a seller $s_i$ and a seller $s_j$ is

\begin{align*}
    \psi(s_i) - \psi(s_j) &= \frac{1}{N-1}\sum_{S \subseteq D \setminus \{s_i, s_j\}} \frac{\mathcal{U}(S \cup \{s_i\}) - \mathcal{U}(S \cup \{s_j\})}{{N-2 \choose |S|}}\\
    &=\text{E}[(\beta_i - \beta_j)\mathcal{U}(\beta_1, \ldots, \beta_N)],
\end{align*}
where $\mathcal{U}(\beta_1, \ldots, \beta_N)$ is the utility evaluated on the appearing sellers and $D$ are all sellers. The Shapley value of sellers can be derived from the estimated Shapley differences between all datum pairs by solving a feasibility problem.
They demonstrate that the algorithm returns an $(\epsilon, \delta)$-approximation with $O(N(logN)^2)$ utility evaluations. The second algorithm is based on their observation that Shapley values are approximately sparse, that is, most values are around the mean.
Exploiting this property, they apply the idea of sparse signal recovering in compressive sensing~\citep{rauhut2010compressive}, and develop an algorithm that produces an $(\epsilon, \delta)$-approximation with only $O(N \log(\log(N )))$ utility evaluations.


\citet{jia2019efficient} further discover that Shapley values for data points used in unweighted kNN classifiers can be computed exactly only in $O(N \log N)$ time. 
Given a testing point $x_{test}$ with label $y_{test}$, they define the utility of a kNN classifier as the likelihood of $y_{test}$, that is,
\begin{equation*}
    \mathcal{U}(S) = \frac{1}{k} \sum_{i=1}^{\text{min}(k, |S|)} \mathbf{1}(y_{\alpha_{i}(S)} = y_{test}),
\end{equation*}
where $\alpha_i(S)$ is the index of the training data that is the $i$-th closest to $x_{test}$ in the set of data points $S$. The special utility function enables efficient computation of Shapley differences between two data points $x_{\alpha_{i}(S)}$ and $x_{\alpha_{i+1}(S)}$, that is, 
\begin{equation}
    \psi(x_{\alpha_{i}(S)}) - \psi(x_{\alpha_{i+1}(S)}) = \frac{\mathbf{1}(y_{\alpha_{i}(S)} = y_{test}) - \mathbf{1}(y_{\alpha_{i+1}(S)} = y_{test})}{k} \frac{\text{min}(i, k)}{i}.
    \label{eq:knn_shapley}
\end{equation}
They start by computing $\psi(x_{\alpha_N(S)}) = \frac{\mathbf{1}(y_{\alpha_{N}(S)} = y_{test})}{N}$ and then exploiting Equation~\ref{eq:knn_shapley} to recursively compute the Shapley values in the order of $x_{\alpha_N(S)}, \ldots, x_{\alpha_1(S)}$.  They further develop an $(\epsilon, \delta)$-approximation algorithm based on Locality Sensitive Hashing~\citep{DBLP:conf/compgeom/DatarIIM04} with only sublinear complexity. The major idea is to only compute Shapley values for the retrieved $k^{*}=\text{max}(k, \frac{1}{\epsilon})$ nearest neighbors of $x_{test}$ and ignore the rest data points, as their Shapley values are too small.
Moreover, they present a Monte-Carlo approximation algorithm with $O(\frac{N}{\epsilon^2}\log(k)\log(\frac{k}{\delta}))$ time complexity for weighted kNN classifiers.


The aforementioned studies~\citep{jia2019efficient, jia2019towards, ghorbani2019data} evaluate the utility of a model by its performance on a validation data set. 
\citet{sim2020collaborative} consider the situation where no validation data sets are available, and propose to use information gain on model parameters as the utility function. Denote by $\theta$ the model parameters. After training on data $D$, the information gain $\text{IG}(\theta)=H(\theta) - H(\theta|D)$ is the reduction in the uncertainty of $\theta$, where $H(\cdot)$ is the entropy function. In addition to Shapley fairness, three additional incentive conditions for revenue allocation are proposed, namely individual rationality, stability of the grand coalition, and group welfare. They also present $p$-Shapley fairness, which assigns a reward $\pi(s_i) = k\psi(s_i)^p$ to a seller $s_i$. By tuning parameter $p \in [0, 1]$, they can trade off between achieving different incentive conditions. Rather than monetary incentives, each participant receives a machine learning model as a reward. To realize different levels of rewards, the models are trained by injecting different levels of noise into training labels.

Federated Learning~\citep{FED_GOOGLE, DBLP:conf/aistats/McMahanMRHA17} enables multiple decentralized participants to collaboratively train a machine learning model while keeping their training data locally.
The data sets contributed by the participants are used in a sequential order determined by a central server.
Evaluating participants' contributions using Shapley value incurs high communication costs among the decentralized participants. Moreover, Shapley value neglects the order of data sources. To accommodate the challenges, \citet{DBLP:series/lncs/0013RZJS20} propose federated Shapley value.
Denote by $\mathcal{U}(s_i + s_j)$ the utility of the model, which is trained on $s_i$'s data first, then on $s_j$'s data. Let $I_{t}$ be the set of selected participants in round $t$ of the federated learning process. The federated Shapley value of participant $s_i$ at round $t$ is defined as follows.

\begin{equation}
\psi_t(s_i) = 
\begin{cases}
    \frac{1}{|I_t|}\sum_{S \subseteq I_t \setminus \{s_i\}}\frac{\mathcal{U}(I_{1:t-1} + (S \cup \{s_i\})) - \mathcal{U}(I_{1:t-1} + S)}{{|I_t| -1 \choose |S|}} & \text{if $s_i \in I_{t}$}\\
    0 & \text{if $s_i \notin I_{t}$}\\
\end{cases}
\label{eq:fed_shapley}
\end{equation}
The federated Shapley value of $s_i$ is $\psi(s_i) = \sum_{t=1}^T\psi(s_i)$, where $T$ is the total rounds in federated learning. The authors show that federated Shapley values satisfy the balance and additivity axioms of Shapley fairness. The other two axioms, symmetry and zero element, are satisfied in each round. They extend the permutation sampling and group testing approximation methods~\citep{jia2019towards} to compute federated Shapley values.


Participants in federated learning spend some costs for contributing their data sets, such as privacy cost~\cite{DBLP:conf/globecom/HuG20} and energy costs~\cite{DBLP:journals/iotj/KangXNXZ19}.
\citet{DBLP:journals/expert/YuLLCCWNY20} propose a fair revenue allocation mechanism for federated learning that jointly considers the costs and contributions of participants. At round $t$, each participant $s_i$ has a public cost $c_i(t)$ and receives a reward $\pi_i(t)$.
The regret $r_i(t)$ of $s_i$ is a function of the difference between the total cost and total reward of $s_i$. A large value of $r_i(t)$ indicates that $s_i$ is not well compensated for the costs incurred to $s_i$.
The authors argue that the payments of participants at each round should achieve contribution fairness and regret fairness.
Contribution fairness requires that the payment $\pi_i(t)$ and the Shapley value $\psi_t(s_i)$ of each participant $s_i$ should be positively correlated, that is, $\sum_{i} \pi_{i}(t)\psi_t(s_i)$ should be maximized. Regret fairness requires that the participants should have similar regrets, that is, the difference of the regrets among participants should be minimized. 
The payments of participants are determined by solving an optimization problem with respect to a budget constraint. Theoretically, they show that the time-averaged regret of participants is upper-bounded by a constant value as $t \rightarrow \infty$.



Shapley value is vulnerable to data-replication attacks. A data provider may replicate his/her data with zero cost and acts as an additional provider to get extra unconscionable rewards.
\citet{agarwal2019marketplace} address the issue by penalizing similar data sets to disincentivize replication, that is, the replication-robust Shapley value is defined as 
\begin{equation*}
    \psi_r(s_i) = \psi(s_i)e^{-\lambda\sum_{s_j \in D \setminus \{s_i\}}\text{SM}(s_i, s_j)},
\end{equation*}
where SM is a similarity metric and $\lambda$ is a constant. However, the proposed replication-robust Shapley value no longer satisfies the balance axiom in Shapley fairness.


\citet{DBLP:journals/corr/abs-2006-14583} study the replication attack in data markets with submodular utility functions. They show that the total reward received by an attacker increases monotonically with respect to the number of the attacker's replications. They discover that the extra reward to the attacker mainly comes from the marginal contributions to small seller groups by the attacker's replication. To fix the issue, the authors propose to down-weigh those contributions when computing Shapley values. Their method guarantees that attackers receive smaller rewards with more replications. 


\citet{DBLP:journals/corr/abs-1911-09052} design a replication robust collaborative data market, where each participant is asked to pay a participation fee. This method discourages replication, as the extra reward received by an attacker cannot cover the attacker's participation cost.

\subsection{Other Revenue Allocation Methods}

There are some other revenue allocation methods in collaborative machine learning other than Shapley value.

Leave-one-out~\citep{doi:10.1080/00401706.1980.10486199} is a commonly used method to evaluate data importance.
It compares the performance of a model trained on the full data set with the performance trained on the full set minus one point. The performance drop is defined as the value of the data point, that is, $\pi(s_i) = \mathcal{U}(D) - \mathcal{U}( D \setminus \{s_i\})$.
Leave-one-out is often approximated by influence function~\citep{doi:10.1080/00401706.1980.10486199, DBLP:conf/icml/KohL17}, which measures how the model changes as the weight of a training point is changed without retraining the model. 
\citet{DBLP:series/lncs/RichardsonFF20} apply an influence function to reward participants in federated learning for their contributed data points. It is shown that the pricing model is incentive compatible. Applying influence functions to price data points are also investigated in~\citep{ richardson2019rewarding, jia2019towards}.
Comparing with Shapley value, leave-one-out methods, in general, are more efficient as they do not require model retraining. However, leave-one-out methods may not accurately assess the values of data points. The methods may assign a low value to one of the two exactly equivalent data points, regardless of how important the datum is, as high performance may still  be achieved by including the other datum~\citep{DBLP:conf/icml/YoonAP20}. 

\citet{DBLP:conf/aaai/YanP21} design a data pricing model based on \textit{core}~\citep{gillies1959solutions}, which is a celebrated revenue allocation solution in cooperative game theory. 
The solution seeks to achieve maximum stability of how participants team up with each other. Core requires that the total reward of each coalition $S$ should be at least equal to the utility $\mathcal{U}(S)$, that is, $\forall S \subseteq D, \sum_{s_i \in S} \pi(s_i) \geq \mathcal{U}(S)$, where $\pi(s_i)$ is the reward of participant $s_i$ and $D$ is the set of all participants. When such a reward cannot be achieved, \textit{least core} relaxes the constraints by allowing a minimum difference $\epsilon$ between the utility of $S$ and the total reward of $S$. In particular, least core computes the payment to each participant by solving the following linear program.
\begin{align}
\begin{split}
     \text{min}\ &\epsilon \\
      \text{s.t. } &\sum_{s_i \in D} \pi(s_i) = \mathcal{U}(D), \\
       &\sum_{s_i \in S} \pi(s_i) + \epsilon \geq \mathcal{U}(S)\ \forall S \subseteq D.
       \label{eq:core}
\end{split}
\end{align}
The number of constraints in Equation~\ref{eq:core} grows exponentially with respect to the number of participants.
\citet{DBLP:conf/aaai/YanP21} tackle the efficiency issue by proposing a Monte Carlo approximation algorithm with guaranteed approximation errors. Their approximation method samples a relatively small number of coalitions and solves Equation~\ref{eq:core} on the sampled coalitions. If Equation~\ref{eq:core} has multiple solutions, the solution with the smallest $l_2$-norm is chosen. Their revenue allocation satisfies the balance, symmetry, and zero element axioms of Shapley fairness.

\citet{DBLP:conf/icml/YoonAP20} propose a reinforcement learning algorithm to value data points. They learn a data value estimator that estimates data values and selects the most valuable samples to train a target classifier. They jointly learn the data value estimator and the corresponding classifier, which enables the classifier and the data value estimator to improve the performance of each other. However, this method cannot guarantee fair revenue distribution among participants. 

Most of the existing revenue allocation methods are developed in the settings that supervised machine learning models are jointly trained. The participants are rewarded based on the contributions of their data sets to the utility of the jointly trained machine learning model. 
To adapt existing pricing models to scenarios where unsupervised machine learning models are jointly trained, the major challenge is to develop a utility function that participants can all agree on.
For some traditional unsupervised machine learning models, there are some widely accepted performance metrics that can serve as the utility functions. For example, Silhouette Coefficient~\cite{rousseeuw1987silhouettes} and Calinski-Harabasz index~\cite{calinski1974dendrite} are widely used to evaluate the performance of clustering algorithms when ground-truth clusters are unknown. However, developing a utility function for some unsupervised models, such as pre-trained deep language models~\cite{DBLP:conf/naacl/DevlinCLT19, DBLP:conf/nips/BrownMRSKDNSSAA20},  may be challenging, as they are evaluated differently in many down-stream machine learning tasks.

\bigskip 

In this section, we review pricing models in collaborative training of machine learning models. The major idea is to price each participant's data set based on its contribution to the performance of the jointly trained machine learning model. Shapley value-based methods guarantee fair revenue distribution among participants, but suffer from poor computational efficiency and scalability. Some alternative methods~\citep{DBLP:conf/icml/YoonAP20, DBLP:conf/aaai/YanP21} enjoy better efficiency or coalition stability, but lose fairness guarantee.

\section{Pricing Machine Learning Models}
\label{sec:price_model}

Machine learning models are needed in many different applications and scenarios.  Rather than building machine learning models from scratch, many users and companies turn to purchase well-trained machine learning models, due to their lack of expertise and computation resources~\citep{DBLP:conf/ndss/YuYZTHJ20, chen2019towards}. 
In this section, we review pricing models for machine learning models and discuss the differences between pricing machine learning models and raw data sets.

\subsection{Pricing Models}
Pricing machine learning models is an emerging research area. To the best of our knowledge, the existing studies mainly focus on arbitrage-free and revenue maximization pricing.

\nop{
\todo{Yong: There is a difference between Price Models and Models' API. For model's API, the inference cost should be considered. This cost may be also varied by different models for the same task. It seems 6.1 only discuss pricing the whole model. I think the inference cost should be considered when pricing a model API, especially those supervised models.}
\resp{Zicun: \citet{agarwal2019marketplace} study the problem of pricing machine learning predictions. As pointed out by Yong, this study does not consider the computational costs of performing predictions. I discuss this limitation at the end of the paragraph of \cite{agarwal2019marketplace}.}
}

\citet{chen2019towards} propose an arbitrage-free and revenue maximization machine learning model marketplace. In their setting, a model owner sells multiple versions of a machine learning model to different buyers. The seller first trains an optimal model on the whole raw data set. 
Then, the seller produces different versions of the optimal model by adding Gaussian noises with different variances to the parameters of the optimal model.
The expected error rates of the generated model instances are monotonically increasing with respect to the variance of the injected noise. An arbitrage-free pricing function guarantees that a buyer cannot derive a high performance model by paying less.
Under their mechanism, a pricing function is arbitrage-free if and only if the function is monotone and subadditive with respect to the inverse of the noise variance. Unfortunately, their pricing model only works for machine learning models trained with strictly convex objective functions.

\citet{chen2019towards} further study revenue maximization in pricing machine learning models with respect to the demands and valuations of a set of buyers. They show that determining the optimal prices is coNP-hard. To overcome the computational hardness, they relax the subadditive constraints $\pi(x+y) \leq \pi(x) + \pi(y)$ by $ \frac{\widehat{\pi}(x)}{x} \leq \frac{\widehat{\pi}(y)}{y}$, where $x \leq y$ and $\widehat{\pi}$ is an approximation of the optimal pricing function $\pi$. They show that $\widehat{\pi}$ is  arbitrage-free and $\forall x > 0$, $\pi(x) / 2 \leq \widehat{\pi}(x) \leq \pi(x)$. They propose a dynamic programming algorithm to compute $\widehat{\pi}$ in $O(n^2)$ time, where $n$ is the number of model versions.

\citet{liu2020dealer} present an end-to-end model marketplace, which jointly considers data owners' privacy costs and model buyers' demands.  A broker collects data from data owners, and produces multiple versions of a machine learning model for sale with different subsets of training data and different differential privacy levels $\epsilon$. The revenues are fully distributed to data owners. Objective perturbation~\citep{DBLP:journals/jmlr/ChaudhuriMS11} is used to train models with required differential privacy levels, which injects quantified random noise into the objective function of a model.
Each data owner $s_i$ requests a minimum compensation for using the owner's data to train a model with $\epsilon$-differential privacy, that is
\begin{equation*}
    \pi(s_i, \epsilon) = b_i \cdot c_i(\epsilon),
\end{equation*}
where $b_i$ is proportional to the Shapley value of $s_i$ with regard to all sellers' data sets and $c_i(\epsilon)$ is the privacy cost of $s_i$. A desirable pricing model should guarantee revenue maximization, arbitrage-freeness with respect to differential privacy levels, and covers the compensations to data owners.  Computing the optimal pricing function is coNP-hard, and thus they propose a dynamic programming algorithm to solve the problem approximately. A limitation of the pricing model is that it cannot adjust prices with respect to dynamic customer demands, which may limit the broker's revenue.

\citet{agarwal2019marketplace} consider an online auction for machine learning model market, which is truthful and revenue maximizing.
They assume buyers come one at a time, and each wants to purchase a machine learning model for the buyer's prediction task. 
Denote by $\mathcal{G}(\widehat{Y}_i, Y_i)$ the quality of a model's prediction $\widehat{Y}_i$ on buyer $i$'s validation data set $Y_i$.
The reward that buyer $i$ receives from the model is $\mu_i \cdot \mathcal{G}(\widehat{Y}_i, Y_i)$, where $\mu_i$ is buyer $i$'s private valuation on unit performance. 
Denote by $p_i$ and $b_i$, respectively, the asking price of the broker and the bid of buyer $i$ for unit performance.
The broker produces a noisy machine learning model for buyer $i$ based on the price difference $p_i - b_i$. Specifically, the model is trained on a data set with quantified injected random noise, such that the model's performance is degraded proportionally to $p_i - b_i$.
Buyer $i$ is charged by a function $RF(p_i, b_i, Y_i)$, which is designed following Myerson's payment function rule~\citep{myerson1981optimal}. The utility that buyer $i$ receives by bidding $b_i$ is
\begin{equation*}
    \mathcal{U}(b_i)=\mu_i \cdot \mathcal{G}(\widehat{Y}_i, Y_i)- RF(p_i,b_i,Y_i),
\end{equation*}
where $\widehat{Y}_i$ is the prediction of the returned noisy model. It is shown that truthfully bidding the buyer's valuation $\mu_i$ can maximize buyer $i$'s utility.
The authors apply a Multiplicative Weights method~\citep{DBLP:journals/toc/AroraHK12} to compute the price $p_i$ from historical revenues. They show that the pricing mechanism achieves maximum revenue.

\subsection{Pricing Raw Data Products Versus Machine Learning Models}

At a high level, pricing machine learning models and raw data sets share a series of common desiderata and techniques.
But their pricing models are essentially different from each other on at least four aspects.

First, the pricing units of machine learning models are often well defined and fixed. A machine learning model is usually priced and sold as a whole. Customers can purchase either a machine learning model or the usage of a machine learning model via API calls, where each call has a fixed price.
In contrast, a raw data set can be consumed in multiple granularities. For example, a customer may be interested in the sales information of American customers in the last year. Another customer, however, may want to purchase the sales information during the Christmas season. Such flexibility makes it easier to version raw data products, and enables more flexible pricing mechanisms. For example, according to how much information is revealed, different prices can be assigned to different queries on the same database~\citep{DBLP:conf/sigmod/DeepK17}. 

Second, versioning in model markets is harder than that in data markets. As data sets have strong and flexible aggregateability, different versions of a data set can be easily produced by aggregating along different dimensions. Producing different versions of a machine learning model requires more sophisticated techniques~\citep{chen2019towards}, since it is challenging to accurately control the differences between multiple versions.

Third, the value of raw data sets to customers is generally harder to measure than that of machine learning models. Often, raw data sets are used to train machine learning models. The ultimate value of a data set dependents not only on its intrinsic properties but also on the specific task that the data set is used for and the analyzing methods~\citep{10.14778/3407790.3407800}. Therefore, it  is usually hard for customers to understand the value of a data set. Many machine learning models are designed for specific tasks and are directly used by people to support decision making~\citep{agarwal2019marketplace}. It is easier for people to verify and understand the value of such machine learning models. For example, customers can value a classification model based on its prediction accuracy.

Last, preventing arbitrage is usually harder in model market than in raw data market. As shown by~\citet{DBLP:conf/uss/TramerZJRR16, DBLP:conf/ndss/YuYZTHJ20}, machine learning models may be stolen by adversaries via a reasonable number of API calls. A customer with a large number of query instances may first purchase some predictions from a target machine learning model. Then, the customer can train a local model with near-equivalent outputs as the target model and use the local model to predict the remaining query instances with almost no cost.

\bigskip

In this section, we review pricing machine learning models. We first revisit arbitrage-free and revenue maximization pricing models. Then, we discuss several major differences between machine learning model products and raw data set products, including pricing units, versioning, arbitrage prevention, and customer valuation.

\section{Conclusions and Future Directions}
\label{sec:conclusion}

In this paper, we survey data pricing in end-to-end machine learning pipelines. We consider three important steps in machine learning pipelines where pricing may be substantially involved, namely raw data collection and labeling, collaborative training machine learning models, and machine learning model marketplaces. We systematically review representative studies in those steps, discuss the pricing principles and review the existing methods. End-to-end machine learning pipelines are playing a more and more important role in the current big data and AI economics era. To the best of our knowledge, this is the first survey on data pricing in  machine learning pipelines.

Data pricing is still in its early stage. There are many research challenges for future works. We list some of them here.

First, the existing studies focus on designing proper rewarding models in each separate stage of machine learning pipelines. There is a lack of systematic study of an end-to-end revenue allocation solution. 
As presented in our survey, the manufacturing process of machine learning models involves multiple parties, including data owners, data processors, machine learning model designers, and other possible participants. Each party provides value-added contributions at one stage of the pipeline and receives a reward. A natural question is how to allocate manufacturing budgets among different parties. 
To answer the question, we need a mechanism to measure and compare the contributions of different parties in different stages. 
We also need a system that can dynamically adjust the budget allocations in response to the changes in supply and demand.

Second, almost all pricing models of collaborative model training formulate revenue allocation as a cooperative game, and use Shapley value to carry out the allocation. They justify the usage of Shapley value through the four axioms, namely balance, symmetry, zero element, and additivity. However, \citet{DBLP:conf/aaai/YanP21} argue that the necessity of additivity for data valuation is debatable. Except for the additivity axiom, many other celebrated allocation solutions in cooperative game theory can also satisfy the other three axioms. Comparing with Shapley value, the other solutions have their advantages and limitations. For example, normalized Banzhaf value~\citep{DBLP:series/synthesis/2011Chalkiadakis} computes the payment to each player as the player's average marginal contribution towards all coalitions of other players. Even though normalized Banzhaf value does not satisfy the additivity axiom, it is more robust to data replication attacks than Shapley value~\citep{DBLP:journals/corr/abs-2006-14583}. 
In a marketplace where robustness is more important than additivity, normalized Banzhaf value is more preferable than Shapley value.
Different types of data marketplaces may have different goals~\citep{10.14778/3407790.3407800}, and thus require different axioms. Therefore, we need a better understanding about the necessary axioms in different marketplaces and explore revenue allocation solutions in specific marketplaces.

Third, fine-grained data procurement for machine learning tasks is not fully explored. In practice, data sets from two sellers may have similar or overlapping parts. A data buyer with a limited budget may not want to purchase many similar data points, as the diversity of training data sets is critical to the performance of machine learning models~\citep{10.14778/3407790.3407800}.
Query-based pricing models~\citep{DBLP:conf/pods/KoutrisUBHS12} allow data buyers to only purchase their interested parts of a data set. However, the existing query-based pricing models are only designed for relational data sets in monopoly markets. Supporting query-based pricing in marketplaces of general data sets with competing sellers brings new challenges and opportunities. For example, it is interesting for data buyers to explore how to distribute their budgets among data sellers to maximize the utility of purchased data sets. For data sellers, it is important to assign prices to different parts of their data sets based on supply and demand, such that the data sellers and their data sets can remain competitive in the market.

Last, rigorous evaluation methods for data pricing models need to be developed. Many existing pricing models are only evaluated in oversimplified experimental environments, where many assumptions are made on the behaviors of market participants. A theoretically sound model, however, may not work in practice, as some model assumptions may break. For example, in a real-world market, participants can have adversarial, ignorant, or coalition-building behaviors. However, the effects of those behaviors on the performance of pricing models are largely dismissed in detailed analysis.
Therefore, as suggested by~\citet{10.14778/3407790.3407800}, a simulation platform that can simulate different behaviors of market participants should be developed. The platform can help us study the advantages and limitations of pricing models in target environments, and choose the best one to deploy.


%
%

\bibliographystyle{spbasic}      
\bibliography{ref}   

\begin{thebibliography}{114}
\providecommand{\natexlab}[1]{#1}
\providecommand{\url}[1]{{#1}}
\providecommand{\urlprefix}{URL }
\expandafter\ifx\csname urlstyle\endcsname\relax
  \providecommand{\doi}[1]{DOI~\discretionary{}{}{}#1}\else
  \providecommand{\doi}{DOI~\discretionary{}{}{}\begingroup
  \urlstyle{rm}\Url}\fi
\providecommand{\eprint}[2][]{\url{#2}}

\bibitem[{Agarwal et~al.(2019)Agarwal, Dahleh, and
  Sarkar}]{agarwal2019marketplace}
Agarwal A, Dahleh MA, Sarkar T (2019) A marketplace for data: An algorithmic
  solution. In: Karlin A, Immorlica N, Johari R (eds) Proceedings of the 2019
  {ACM} Conference on Economics and Computation, {EC} 2019, Phoenix, AZ, USA,
  June 24-28, 2019, {ACM}, pp 701--726, \doi{10.1145/3328526.3329589},
  \urlprefix\url{https://doi.org/10.1145/3328526.3329589}

\bibitem[{de~Alfaro et~al.(2016)de~Alfaro, Faella, Polychronopoulos, and
  Shavlovsky}]{de2016incentives}
de~Alfaro L, Faella M, Polychronopoulos V, Shavlovsky M (2016) Incentives for
  truthful evaluations. CoRR abs/1608.07886,
  \urlprefix\url{http://arxiv.org/abs/1608.07886}, \eprint{1608.07886}

\bibitem[{Arora et~al.(2012)Arora, Hazan, and
  Kale}]{DBLP:journals/toc/AroraHK12}
Arora S, Hazan E, Kale S (2012) The multiplicative weights update method: a
  meta-algorithm and applications. Theory Comput 8(1):121--164,
  \doi{10.4086/toc.2012.v008a006},
  \urlprefix\url{https://doi.org/10.4086/toc.2012.v008a006}

\bibitem[{Ausubel et~al.(2006)Ausubel, Milgrom et~al.}]{ausubel2006lovely}
Ausubel LM, Milgrom P, et~al. (2006) The lovely but lonely vickrey auction.
  Combinatorial auctions 17:22--26

\bibitem[{Balasubramanian et~al.(2015)Balasubramanian, Bhattacharya, and
  Krishnan}]{balasubramanian2015pricing}
Balasubramanian S, Bhattacharya S, Krishnan VV (2015) Pricing information
  goods: A strategic analysis of the selling and pay-per-use mechanisms.
  Marketing Science 34(2):218--234

\bibitem[{BDEX(2021)}]{BDEX}
BDEX (2021) Bdex. \url{https://www.bdex.com}, accessed: 2021-05-09

\bibitem[{Brendan and Daniel(2017)}]{FED_GOOGLE}
Brendan M, Daniel R (2017) Federated learning: Collaborative machine learning
  without centralized training data. Google AI Blog
  \urlprefix\url{https://ai.googleblog.com/2017/04/federated-learning-collaborative.html},
  accessed: 2021-07-02

\bibitem[{Brennan et~al.(2013)Brennan, Canning, and Mcdowell}]{Brennan13}
Brennan R, Canning L, Mcdowell R (2013) Business-to-business marketing. Sage
  Publications, \doi{10.4135/9781446276518}

\bibitem[{Brown et~al.(2020)Brown, Mann, Ryder, Subbiah, Kaplan, Dhariwal,
  Neelakantan, Shyam, Sastry, Askell, Agarwal, Herbert{-}Voss, Krueger,
  Henighan, Child, Ramesh, Ziegler, Wu, Winter, Hesse, Chen, Sigler, Litwin,
  Gray, Chess, Clark, Berner, McCandlish, Radford, Sutskever, and
  Amodei}]{DBLP:conf/nips/BrownMRSKDNSSAA20}
Brown TB, Mann B, Ryder N, Subbiah M, Kaplan J, Dhariwal P, Neelakantan A,
  Shyam P, Sastry G, Askell A, Agarwal S, Herbert{-}Voss A, Krueger G, Henighan
  T, Child R, Ramesh A, Ziegler DM, Wu J, Winter C, Hesse C, Chen M, Sigler E,
  Litwin M, Gray S, Chess B, Clark J, Berner C, McCandlish S, Radford A,
  Sutskever I, Amodei D (2020) Language models are few-shot learners. In:
  Larochelle H, Ranzato M, Hadsell R, Balcan M, Lin H (eds) Advances in Neural
  Information Processing Systems 33: Annual Conference on Neural Information
  Processing Systems 2020, NeurIPS 2020, December 6-12, 2020, virtual,
  \urlprefix\url{https://proceedings.neurips.cc/paper/2020/hash/1457c0d6bfcb4967418bfb8ac142f64a-Abstract.html}

\bibitem[{Buneman and Tan(2007)}]{DBLP:conf/sigmod/BunemanT07}
Buneman P, Tan WC (2007) Provenance in databases. In: Chan CY, Ooi BC, Zhou A
  (eds) Proceedings of the {ACM} {SIGMOD} International Conference on
  Management of Data, Beijing, China, June 12-14, 2007, {ACM}, pp 1171--1173,
  \doi{10.1145/1247480.1247646},
  \urlprefix\url{https://doi.org/10.1145/1247480.1247646}

\bibitem[{Burkett(2006)}]{BurkettJohnP2006MOEa}
Burkett JP (2006) Microeconomics: Optimization, Experiments, and Behavior. OUP
  Catalogue, Oxford University Press, New York

\bibitem[{Cali{\'n}ski and Harabasz(1974)}]{calinski1974dendrite}
Cali{\'n}ski T, Harabasz J (1974) A dendrite method for cluster analysis.
  Communications in Statistics-theory and Methods 3(1):1--27

\bibitem[{Chalkiadakis et~al.(2011)Chalkiadakis, Elkind, and
  Wooldridge}]{DBLP:series/synthesis/2011Chalkiadakis}
Chalkiadakis G, Elkind E, Wooldridge MJ (2011) Computational Aspects of
  Cooperative Game Theory. Synthesis Lectures on Artificial Intelligence and
  Machine Learning, Morgan {\&} Claypool Publishers,
  \doi{10.2200/S00355ED1V01Y201107AIM016},
  \urlprefix\url{https://doi.org/10.2200/S00355ED1V01Y201107AIM016}

\bibitem[{Chaudhuri et~al.(2011)Chaudhuri, Monteleoni, and
  Sarwate}]{DBLP:journals/jmlr/ChaudhuriMS11}
Chaudhuri K, Monteleoni C, Sarwate AD (2011) Differentially private empirical
  risk minimization. J Mach Learn Res 12:1069--1109,
  \urlprefix\url{http://dl.acm.org/citation.cfm?id=2021036}

\bibitem[{Chawla et~al.(2019)Chawla, Deep, Koutris, and
  Teng}]{DBLP:journals/pvldb/ChawlaDKT19}
Chawla S, Deep S, Koutris P, Teng Y (2019) Revenue maximization for query
  pricing. Proc {VLDB} Endow 13(1):1--14, \doi{10.14778/3357377.3357378},
  \urlprefix\url{http://www.vldb.org/pvldb/vol13/p1-chawla.pdf}

\bibitem[{Chen et~al.(2019)Chen, Koutris, and Kumar}]{chen2019towards}
Chen L, Koutris P, Kumar A (2019) Towards model-based pricing for machine
  learning in a data marketplace. In: Boncz PA, Manegold S, Ailamaki A,
  Deshpande A, Kraska T (eds) Proceedings of the 2019 International Conference
  on Management of Data, {SIGMOD} Conference 2019, Amsterdam, The Netherlands,
  June 30 - July 5, 2019, {ACM}, pp 1535--1552, \doi{10.1145/3299869.3300078},
  \urlprefix\url{https://doi.org/10.1145/3299869.3300078}

\bibitem[{Chen et~al.(2020)Chen, Zaharia, and Zou}]{chen2020frugalml}
Chen L, Zaharia M, Zou JY (2020) Frugalml: How to use ml prediction apis more
  accurately and cheaply. Advances in Neural Information Processing Systems 33

\bibitem[{Cook and Weisberg(1980)}]{doi:10.1080/00401706.1980.10486199}
Cook RD, Weisberg S (1980) Characterizations of an empirical influence function
  for detecting influential cases in regression. Technometrics 22(4):495--508,
  \doi{10.1080/00401706.1980.10486199},
  \urlprefix\url{https://www.tandfonline.com/doi/abs/10.1080/00401706.1980.10486199},
  \eprint{https://www.tandfonline.com/doi/pdf/10.1080/00401706.1980.10486199}

\bibitem[{Dandekar et~al.(2012)Dandekar, Fawaz, and
  Ioannidis}]{DBLP:conf/wine/DandekarFI12}
Dandekar P, Fawaz N, Ioannidis S (2012) Privacy auctions for recommender
  systems. In: Goldberg PW (ed) Internet and Network Economics - 8th
  International Workshop, {WINE} 2012, Liverpool, UK, December 10-12, 2012.
  Proceedings, Springer, Lecture Notes in Computer Science, vol 7695, pp
  309--322, \doi{10.1007/978-3-642-35311-6\_23},
  \urlprefix\url{https://doi.org/10.1007/978-3-642-35311-6\_23}

\bibitem[{Dasgupta and Ghosh(2013)}]{DBLP:conf/www/DasguptaG13}
Dasgupta A, Ghosh A (2013) Crowdsourced judgement elicitation with endogenous
  proficiency. In: Schwabe D, Almeida VAF, Glaser H, Baeza{-}Yates R, Moon SB
  (eds) 22nd International World Wide Web Conference, {WWW} '13, Rio de
  Janeiro, Brazil, May 13-17, 2013, International World Wide Web Conferences
  Steering Committee / {ACM}, pp 319--330, \doi{10.1145/2488388.2488417},
  \urlprefix\url{https://doi.org/10.1145/2488388.2488417}

\bibitem[{Datar et~al.(2004)Datar, Immorlica, Indyk, and
  Mirrokni}]{DBLP:conf/compgeom/DatarIIM04}
Datar M, Immorlica N, Indyk P, Mirrokni VS (2004) Locality-sensitive hashing
  scheme based on p-stable distributions. In: Snoeyink J, Boissonnat J (eds)
  Proceedings of the 20th {ACM} Symposium on Computational Geometry, Brooklyn,
  New York, USA, June 8-11, 2004, {ACM}, pp 253--262,
  \doi{10.1145/997817.997857},
  \urlprefix\url{https://doi.org/10.1145/997817.997857}

\bibitem[{Dawex(2021)}]{DAWEX}
Dawex (2021) Dawex. \url{https://www.dawex.com/en/}, accessed: 2021-05-09

\bibitem[{De~Toni et~al.(2017)De~Toni, Milan, Saciloto, and
  Larentis}]{de2017pricing}
De~Toni D, Milan GS, Saciloto EB, Larentis F (2017) Pricing strategies and
  levels and their impact on corporate profitability. Revista de
  Administra{\c{c}}{\~a}o (S{\~a}o Paulo) 52(2):120--133

\bibitem[{Deep and Koutris(2017{\natexlab{a}})}]{DBLP:conf/icdt/DeepK17}
Deep S, Koutris P (2017{\natexlab{a}}) The design of arbitrage-free data
  pricing schemes. In: Benedikt M, Orsi G (eds) 20th International Conference
  on Database Theory, {ICDT} 2017, March 21-24, 2017, Venice, Italy, Schloss
  Dagstuhl - Leibniz-Zentrum f{\"{u}}r Informatik, LIPIcs, vol~68, pp
  12:1--12:18, \doi{10.4230/LIPIcs.ICDT.2017.12},
  \urlprefix\url{https://doi.org/10.4230/LIPIcs.ICDT.2017.12}

\bibitem[{Deep and Koutris(2017{\natexlab{b}})}]{DBLP:conf/sigmod/DeepK17}
Deep S, Koutris P (2017{\natexlab{b}}) {QIRANA:} {A} framework for scalable
  query pricing. In: Salihoglu S, Zhou W, Chirkova R, Yang J, Suciu D (eds)
  Proceedings of the 2017 {ACM} International Conference on Management of Data,
  {SIGMOD} Conference 2017, Chicago, IL, USA, May 14-19, 2017, {ACM}, pp
  699--713, \doi{10.1145/3035918.3064017},
  \urlprefix\url{https://doi.org/10.1145/3035918.3064017}

\bibitem[{Devlin et~al.(2019)Devlin, Chang, Lee, and
  Toutanova}]{DBLP:conf/naacl/DevlinCLT19}
Devlin J, Chang M, Lee K, Toutanova K (2019) {BERT:} pre-training of deep
  bidirectional transformers for language understanding. In: Burstein J, Doran
  C, Solorio T (eds) Proceedings of the 2019 Conference of the North American
  Chapter of the Association for Computational Linguistics: Human Language
  Technologies, {NAACL-HLT} 2019, Minneapolis, MN, USA, June 2-7, 2019, Volume
  1 (Long and Short Papers), Association for Computational Linguistics, pp
  4171--4186, \doi{10.18653/v1/n19-1423},
  \urlprefix\url{https://doi.org/10.18653/v1/n19-1423}

\bibitem[{Dibb et~al.(2005)Dibb, Simkin, Pride, and Ferrell}]{oro2041}
Dibb S, Simkin L, Pride WM, Ferrell O (2005) Marketing: Concepts and
  Strategies. 5th Edition. Houghton Mifflin, Abingdon, UK

\bibitem[{Dwork(2008)}]{DBLP:conf/tamc/Dwork08}
Dwork C (2008) Differential privacy: {A} survey of results. In: Agrawal M, Du
  D, Duan Z, Li A (eds) Theory and Applications of Models of Computation, 5th
  International Conference, {TAMC} 2008, Xi'an, China, April 25-29, 2008.
  Proceedings, Springer, Lecture Notes in Computer Science, vol 4978, pp 1--19,
  \doi{10.1007/978-3-540-79228-4\_1},
  \urlprefix\url{https://doi.org/10.1007/978-3-540-79228-4\_1}

\bibitem[{Ensthaler and Giebe(2014)}]{DBLP:journals/eor/EnsthalerG14}
Ensthaler L, Giebe T (2014) Bayesian optimal knapsack procurement. Eur J Oper
  Res 234(3):774--779, \doi{10.1016/j.ejor.2013.09.031},
  \urlprefix\url{https://doi.org/10.1016/j.ejor.2013.09.031}

\bibitem[{Fernandez et~al.(2020)Fernandez, Subramaniam, and
  Franklin}]{10.14778/3407790.3407800}
Fernandez RC, Subramaniam P, Franklin MJ (2020) Data market platforms: Trading
  data assets to solve data problems. Proc VLDB Endow 13(12):1933--1947,
  \doi{10.14778/3407790.3407800}

\bibitem[{Fricker and Maksimov(2017)}]{10.1007/978-3-319-69191-6_4}
Fricker SA, Maksimov YV (2017) Pricing of data products in data marketplaces.
  In: Ojala A, Holmstr{\"o}m~Olsson H, Werder K (eds) Software Business,
  Springer International Publishing, Cham, pp 49--66

\bibitem[{Fung and Beschastnikh(2019)}]{fung2019brokered}
Fung C, Beschastnikh I (2019) Brokered agreements in multi-party machine
  learning. In: Proceedings of the 10th ACM SIGOPS Asia-Pacific Workshop on
  Systems, pp 69--75

\bibitem[{Ghorbani and Zou(2019)}]{ghorbani2019data}
Ghorbani A, Zou JY (2019) Data shapley: Equitable valuation of data for machine
  learning. In: Chaudhuri K, Salakhutdinov R (eds) Proceedings of the 36th
  International Conference on Machine Learning, {ICML} 2019, 9-15 June 2019,
  Long Beach, California, {USA}, {PMLR}, Proceedings of Machine Learning
  Research, vol~97, pp 2242--2251,
  \urlprefix\url{http://proceedings.mlr.press/v97/ghorbani19c.html}

\bibitem[{Ghosh and Roth(2011)}]{DBLP:conf/sigecom/GhoshR11}
Ghosh A, Roth A (2011) Selling privacy at auction. In: Shoham Y, Chen Y,
  Roughgarden T (eds) Proceedings 12th {ACM} Conference on Electronic Commerce
  (EC-2011), San Jose, CA, USA, June 5-9, 2011, {ACM}, pp 199--208,
  \doi{10.1145/1993574.1993605},
  \urlprefix\url{https://doi.org/10.1145/1993574.1993605}

\bibitem[{Gillies(1959)}]{gillies1959solutions}
Gillies DB (1959) Solutions to general non-zero-sum games. Contributions to the
  Theory of Games 4:47--85

\bibitem[{Goel and Faltings(2019)}]{DBLP:conf/aaai/GoelF19}
Goel N, Faltings B (2019) Deep bayesian trust: {A} dominant and fair incentive
  mechanism for crowd. In: The Thirty-Third {AAAI} Conference on Artificial
  Intelligence, {AAAI} 2019, The Thirty-First Innovative Applications of
  Artificial Intelligence Conference, {IAAI} 2019, The Ninth {AAAI} Symposium
  on Educational Advances in Artificial Intelligence, {EAAI} 2019, Honolulu,
  Hawaii, USA, January 27 - February 1, 2019, {AAAI} Press, pp 1996--2003,
  \doi{10.1609/aaai.v33i01.33011996},
  \urlprefix\url{https://doi.org/10.1609/aaai.v33i01.33011996}

\bibitem[{Gur et~al.(2014)Gur, Zeevi, and Besbes}]{DBLP:conf/nips/GurZB14}
Gur Y, Zeevi AJ, Besbes O (2014) Stochastic multi-armed-bandit problem with
  non-stationary rewards. In: Ghahramani Z, Welling M, Cortes C, Lawrence ND,
  Weinberger KQ (eds) Advances in Neural Information Processing Systems 27:
  Annual Conference on Neural Information Processing Systems 2014, December
  8-13 2014, Montreal, Quebec, Canada, pp 199--207,
  \urlprefix\url{https://proceedings.neurips.cc/paper/2014/hash/903ce9225fca3e988c2af215d4e544d3-Abstract.html}

\bibitem[{Han et~al.(2020)Han, Tople, Rogers, Wooldridge, Ohrimenko, and
  Tschiatschek}]{DBLP:journals/corr/abs-2006-14583}
Han D, Tople S, Rogers A, Wooldridge MJ, Ohrimenko O, Tschiatschek S (2020)
  Replication-robust payoff-allocation with applications in machine learning
  marketplaces. CoRR abs/2006.14583,
  \urlprefix\url{https://arxiv.org/abs/2006.14583}, \eprint{2006.14583}

\bibitem[{Heckman et~al.(2015)Heckman, Boehmer, Peters, Davaloo, and
  Kurup}]{heckman2015pricing}
Heckman JR, Boehmer EL, Peters EH, Davaloo M, Kurup NG (2015) A pricing model
  for data markets. iConference 2015 Proceedings

\bibitem[{Hoeffding(1994)}]{hoeffding1994probability}
Hoeffding W (1994) Probability inequalities for sums of bounded random
  variables. In: The Collected Works of Wassily Hoeffding, Springer, pp
  409--426

\bibitem[{Hu and Gong(2020)}]{DBLP:conf/globecom/HuG20}
Hu R, Gong Y (2020) Trading data for learning: Incentive mechanism for
  on-device federated learning. In: {IEEE} Global Communications Conference,
  {GLOBECOM} 2020, Virtual Event, Taiwan, December 7-11, 2020, {IEEE}, pp 1--6,
  \doi{10.1109/GLOBECOM42002.2020.9322475},
  \urlprefix\url{https://doi.org/10.1109/GLOBECOM42002.2020.9322475}

\bibitem[{Hu et~al.(2018)Hu, Liang, Zhang, Li, and
  Liu}]{DBLP:conf/nips/HuLZLL18}
Hu Z, Liang Y, Zhang J, Li Z, Liu Y (2018) Inference aided reinforcement
  learning for incentive mechanism design in crowdsourcing. In: Bengio S,
  Wallach HM, Larochelle H, Grauman K, Cesa{-}Bianchi N, Garnett R (eds)
  Advances in Neural Information Processing Systems 31: Annual Conference on
  Neural Information Processing Systems 2018, NeurIPS 2018, December 3-8, 2018,
  Montr{\'{e}}al, Canada, pp 5512--5522,
  \urlprefix\url{https://proceedings.neurips.cc/paper/2018/hash/f2e43fa3400d826df4195a9ac70dca62-Abstract.html}

\bibitem[{Hynes et~al.(2018)Hynes, Dao, Yan, Cheng, and
  Song}]{DBLP:journals/pvldb/HynesDYCS18}
Hynes N, Dao D, Yan D, Cheng R, Song D (2018) A demonstration of sterling: {A}
  privacy-preserving data marketplace. Proc {VLDB} Endow 11(12):2086--2089,
  \doi{10.14778/3229863.3236266},
  \urlprefix\url{http://www.vldb.org/pvldb/vol11/p2086-hynes.pdf}

\bibitem[{Irvin(1978)}]{Irvin79}
Irvin G (1978) Modern Cost-Benefit Methods. Macmillan Publishers Limited,
  London, \doi{10.1007/978-1-349-15912-3}

\bibitem[{Jia et~al.(2019{\natexlab{a}})Jia, Dao, Wang, Hubis, G{\"{u}}rel, Li,
  Zhang, Spanos, and Song}]{jia2019efficient}
Jia R, Dao D, Wang B, Hubis FA, G{\"{u}}rel NM, Li B, Zhang C, Spanos CJ, Song
  D (2019{\natexlab{a}}) Efficient task-specific data valuation for nearest
  neighbor algorithms. Proc {VLDB} Endow 12(11):1610--1623,
  \doi{10.14778/3342263.3342637},
  \urlprefix\url{http://www.vldb.org/pvldb/vol12/p1610-jia.pdf}

\bibitem[{Jia et~al.(2019{\natexlab{b}})Jia, Dao, Wang, Hubis, Hynes,
  G{\"{u}}rel, Li, Zhang, Song, and Spanos}]{jia2019towards}
Jia R, Dao D, Wang B, Hubis FA, Hynes N, G{\"{u}}rel NM, Li B, Zhang C, Song D,
  Spanos CJ (2019{\natexlab{b}}) Towards efficient data valuation based on the
  shapley value. In: Chaudhuri K, Sugiyama M (eds) The 22nd International
  Conference on Artificial Intelligence and Statistics, {AISTATS} 2019, 16-18
  April 2019, Naha, Okinawa, Japan, {PMLR}, Proceedings of Machine Learning
  Research, vol~89, pp 1167--1176,
  \urlprefix\url{http://proceedings.mlr.press/v89/jia19a.html}

\bibitem[{Jiang et~al.(2015)Jiang, Gao, Duan, and Huang}]{jiang2015economics}
Jiang C, Gao L, Duan L, Huang J (2015) Economics of peer-to-peer mobile
  crowdsensing. In: 2015 IEEE Global Communications Conference (GLOBECOM),
  IEEE, pp 1--6

\bibitem[{Jin et~al.(2015)Jin, Su, Chen, Nahrstedt, and Xu}]{jin2015quality}
Jin H, Su L, Chen D, Nahrstedt K, Xu J (2015) Quality of information aware
  incentive mechanisms for mobile crowd sensing systems. In: Shen SX, Sun Y,
  Chen J, Zhang J, Zussman G (eds) Proceedings of the 16th {ACM} International
  Symposium on Mobile Ad Hoc Networking and Computing, MobiHoc 2015, Hangzhou,
  China, June 22-25, 2015, {ACM}, pp 167--176, \doi{10.1145/2746285.2746310},
  \urlprefix\url{https://doi.org/10.1145/2746285.2746310}

\bibitem[{Jin et~al.(2019)Jin, Xiao, Li, and Guo}]{DBLP:conf/infocom/JinXLG19}
Jin W, Xiao M, Li M, Guo L (2019) If you do not care about it, sell it: Trading
  location privacy in mobile crowd sensing. In: 2019 {IEEE} Conference on
  Computer Communications, {INFOCOM} 2019, Paris, France, April 29 - May 2,
  2019, {IEEE}, pp 1045--1053, \doi{10.1109/INFOCOM.2019.8737457},
  \urlprefix\url{https://doi.org/10.1109/INFOCOM.2019.8737457}

\bibitem[{Jorgensen et~al.(2015)Jorgensen, Yu, and
  Cormode}]{DBLP:conf/icde/JorgensenYC15}
Jorgensen Z, Yu T, Cormode G (2015) Conservative or liberal? personalized
  differential privacy. In: Gehrke J, Lehner W, Shim K, Cha SK, Lohman GM (eds)
  31st {IEEE} International Conference on Data Engineering, {ICDE} 2015, Seoul,
  South Korea, April 13-17, 2015, {IEEE} Computer Society, pp 1023--1034,
  \doi{10.1109/ICDE.2015.7113353},
  \urlprefix\url{https://doi.org/10.1109/ICDE.2015.7113353}

\bibitem[{Kang et~al.(2019)Kang, Xiong, Niyato, Xie, and
  Zhang}]{DBLP:journals/iotj/KangXNXZ19}
Kang J, Xiong Z, Niyato D, Xie S, Zhang J (2019) Incentive mechanism for
  reliable federated learning: {A} joint optimization approach to combining
  reputation and contract theory. {IEEE} Internet Things J 6(6):10700--10714,
  \doi{10.1109/JIOT.2019.2940820},
  \urlprefix\url{https://doi.org/10.1109/JIOT.2019.2940820}

\bibitem[{Koh and Liang(2017)}]{DBLP:conf/icml/KohL17}
Koh PW, Liang P (2017) Understanding black-box predictions via influence
  functions. In: Precup D, Teh YW (eds) Proceedings of the 34th International
  Conference on Machine Learning, {ICML} 2017, Sydney, NSW, Australia, 6-11
  August 2017, {PMLR}, Proceedings of Machine Learning Research, vol~70, pp
  1885--1894, \urlprefix\url{http://proceedings.mlr.press/v70/koh17a.html}

\bibitem[{Koutris et~al.(2012)Koutris, Upadhyaya, Balazinska, Howe, and
  Suciu}]{DBLP:conf/pods/KoutrisUBHS12}
Koutris P, Upadhyaya P, Balazinska M, Howe B, Suciu D (2012) Query-based data
  pricing. In: Benedikt M, Kr{\"{o}}tzsch M, Lenzerini M (eds) Proceedings of
  the 31st {ACM} {SIGMOD-SIGACT-SIGART} Symposium on Principles of Database
  Systems, {PODS} 2012, Scottsdale, AZ, USA, May 20-24, 2012, {ACM}, pp
  167--178, \doi{10.1145/2213556.2213582},
  \urlprefix\url{https://doi.org/10.1145/2213556.2213582}

\bibitem[{Koutris et~al.(2013)Koutris, Upadhyaya, Balazinska, Howe, and
  Suciu}]{DBLP:conf/sigmod/KoutrisUBHS13}
Koutris P, Upadhyaya P, Balazinska M, Howe B, Suciu D (2013) Toward practical
  query pricing with querymarket. In: Ross KA, Srivastava D, Papadias D (eds)
  Proceedings of the {ACM} {SIGMOD} International Conference on Management of
  Data, {SIGMOD} 2013, New York, NY, USA, June 22-27, 2013, {ACM}, pp 613--624,
  \doi{10.1145/2463676.2465335},
  \urlprefix\url{https://doi.org/10.1145/2463676.2465335}

\bibitem[{Koutsopoulos(2013)}]{koutsopoulos2013optimal}
Koutsopoulos I (2013) Optimal incentive-driven design of participatory sensing
  systems. In: Proceedings of the {IEEE} {INFOCOM} 2013, Turin, Italy, April
  14-19, 2013, {IEEE}, pp 1402--1410, \doi{10.1109/INFCOM.2013.6566934},
  \urlprefix\url{https://doi.org/10.1109/INFCOM.2013.6566934}

\bibitem[{Leyton{-}Brown and Shoham(2008)}]{DBLP:series/synthesis/2008Leyton}
Leyton{-}Brown K, Shoham Y (2008) Essentials of Game Theory: {A} Concise
  Multidisciplinary Introduction. Synthesis Lectures on Artificial Intelligence
  and Machine Learning, Morgan {\&} Claypool Publishers,
  \doi{10.2200/S00108ED1V01Y200802AIM003},
  \urlprefix\url{https://doi.org/10.2200/S00108ED1V01Y200802AIM003}

\bibitem[{Li et~al.(2013)Li, Li, Miklau, and Suciu}]{DBLP:conf/icdt/LiLMS13}
Li C, Li DY, Miklau G, Suciu D (2013) A theory of pricing private data. In: Tan
  W, Guerrini G, Catania B, Gounaris A (eds) Joint 2013 {EDBT/ICDT}
  Conferences, {ICDT} '13 Proceedings, Genoa, Italy, March 18-22, 2013, {ACM},
  pp 33--44, \doi{10.1145/2448496.2448502},
  \urlprefix\url{https://doi.org/10.1145/2448496.2448502}

\bibitem[{Liang et~al.(2018)Liang, Yu, An, Yang, Fu, and
  Zhao}]{liang2018survey}
Liang F, Yu W, An D, Yang Q, Fu X, Zhao W (2018) A survey on big data market:
  Pricing, trading and protection. IEEE Access 6:15132--15154

\bibitem[{Lin and Kifer(2014)}]{DBLP:journals/pvldb/LinK14}
Lin B, Kifer D (2014) On arbitrage-free pricing for general data queries. Proc
  {VLDB} Endow 7(9):757--768, \doi{10.14778/2732939.2732948},
  \urlprefix\url{http://www.vldb.org/pvldb/vol7/p757-lin.pdf}

\bibitem[{Liu et~al.(2016)Liu, Chakraborty, and
  Mittal}]{DBLP:conf/ndss/LiuMC16}
Liu C, Chakraborty S, Mittal P (2016) Dependence makes you vulnberable:
  Differential privacy under dependent tuples. In: 23rd Annual Network and
  Distributed System Security Symposium, {NDSS} 2016, San Diego, California,
  USA, February 21-24, 2016, The Internet Society,
  \urlprefix\url{http://wp.internetsociety.org/ndss/wp-content/uploads/sites/25/2017/09/dependence-makes-you-vulnerable-differential-privacy-under-dependent-tuples.pdf}

\bibitem[{Liu et~al.(2021)Liu, Lou, Liu, Xiong, Pei, and Sun}]{liu2020dealer}
Liu J, Lou J, Liu J, Xiong L, Pei J, Sun J (2021) Dealer: An end-to-end model
  marketplace with differential privacy. Proc VLDB Endow 14(6):957--969,
  \doi{10.14778/3447689.3447700},
  \urlprefix\url{https://doi.org/10.14778/3447689.3447700}

\bibitem[{Liu and Chen(2017{\natexlab{a}})}]{liu2017machine}
Liu Y, Chen Y (2017{\natexlab{a}}) Machine-learning aided peer prediction. In:
  Daskalakis C, Babaioff M, Moulin H (eds) Proceedings of the 2017 {ACM}
  Conference on Economics and Computation, {EC} '17, Cambridge, MA, USA, June
  26-30, 2017, {ACM}, pp 63--80, \doi{10.1145/3033274.3085126},
  \urlprefix\url{https://doi.org/10.1145/3033274.3085126}

\bibitem[{Liu and Chen(2017{\natexlab{b}})}]{DBLP:conf/aaai/LiuC17}
Liu Y, Chen Y (2017{\natexlab{b}}) Sequential peer prediction: Learning to
  elicit effort using posted prices. In: Singh SP, Markovitch S (eds)
  Proceedings of the Thirty-First {AAAI} Conference on Artificial Intelligence,
  February 4-9, 2017, San Francisco, California, {USA}, {AAAI} Press, pp
  607--613,
  \urlprefix\url{http://aaai.org/ocs/index.php/AAAI/AAAI17/paper/view/14970}

\bibitem[{Louis(2020)}]{forbes}
Louis C (2020) Roundup of machine learning forecasts and market estimates,
  2020. Forbes
  \urlprefix\url{https://www.forbes.com/sites/louiscolumbus/2020/01/19/roundup-of-machine-learning-forecasts-and-market-estimates-2020},
  accessed: 2021-06-28

\bibitem[{Luong et~al.(2016)Luong, Hoang, Wang, Niyato, Kim, and
  Han}]{luong2016data}
Luong NC, Hoang DT, Wang P, Niyato D, Kim DI, Han Z (2016) Data collection and
  wireless communication in internet of things (iot) using economic analysis
  and pricing models: A survey. IEEE Communications Surveys \& Tutorials
  18(4):2546--2590

\bibitem[{Ma et~al.(2020)Ma, Zhang, Wang, Ruan, Wang, Tang, Ma, Gao, and
  Gao}]{DBLP:conf/aaai/MaZWRWTMGG20}
Ma L, Zhang C, Wang Y, Ruan W, Wang J, Tang W, Ma X, Gao X, Gao J (2020)
  Concare: Personalized clinical feature embedding via capturing the healthcare
  context. In: The Thirty-Fourth {AAAI} Conference on Artificial Intelligence,
  {AAAI} 2020, The Thirty-Second Innovative Applications of Artificial
  Intelligence Conference, {IAAI} 2020, The Tenth {AAAI} Symposium on
  Educational Advances in Artificial Intelligence, {EAAI} 2020, New York, NY,
  USA, February 7-12, 2020, {AAAI} Press, pp 833--840,
  \urlprefix\url{https://aaai.org/ojs/index.php/AAAI/article/view/5428}

\bibitem[{Maleki et~al.(2013)Maleki, Tran{-}Thanh, Hines, Rahwan, and
  Rogers}]{maleki2013bounding}
Maleki S, Tran{-}Thanh L, Hines G, Rahwan T, Rogers A (2013) Bounding the
  estimation error of sampling-based shapley value approximation with/without
  stratifying. CoRR abs/1306.4265,
  \urlprefix\url{http://arxiv.org/abs/1306.4265}, \eprint{1306.4265}

\bibitem[{McMahan et~al.(2017)McMahan, Moore, Ramage, Hampson, and
  y~Arcas}]{DBLP:conf/aistats/McMahanMRHA17}
McMahan B, Moore E, Ramage D, Hampson S, y~Arcas BA (2017)
  Communication-efficient learning of deep networks from decentralized data.
  In: Singh A, Zhu XJ (eds) Proceedings of the 20th International Conference on
  Artificial Intelligence and Statistics, {AISTATS} 2017, 20-22 April 2017,
  Fort Lauderdale, FL, {USA}, {PMLR}, Proceedings of Machine Learning Research,
  vol~54, pp 1273--1282,
  \urlprefix\url{http://proceedings.mlr.press/v54/mcmahan17a.html}

\bibitem[{Miao et~al.(2020)Miao, Gao, Chen, Peng, Yin, and
  Li}]{miao2020towards}
Miao X, Gao Y, Chen L, Peng H, Yin J, Li Q (2020) Towards query pricing on
  incomplete data. IEEE Transactions on Knowledge and Data Engineering

\bibitem[{Muschalle et~al.(2012)Muschalle, Stahl, L{\"o}ser, and
  Vossen}]{muschalle2012pricing}
Muschalle A, Stahl F, L{\"o}ser A, Vossen G (2012) Pricing approaches for data
  markets. In: International workshop on business intelligence for the
  real-time enterprise, Springer, pp 129--144

\bibitem[{Myerson(1981)}]{myerson1981optimal}
Myerson RB (1981) Optimal auction design. Mathematics of operations research
  6(1):58--73

\bibitem[{Nagle(2010)}]{2135}
Nagle J TT \&~Hogan (2010) The Strategy and Tactics of Pricing: A Guide to
  Growing More Profitably. Prentice Hall

\bibitem[{Nash et~al.(2007)Nash, Segoufin, and Vianu}]{DBLP:conf/icdt/NashSV07}
Nash A, Segoufin L, Vianu V (2007) Determinacy and rewriting of conjunctive
  queries using views: {A} progress report. In: Schwentick T, Suciu D (eds)
  Database Theory - {ICDT} 2007, 11th International Conference, Barcelona,
  Spain, January 10-12, 2007, Proceedings, Springer, Lecture Notes in Computer
  Science, vol 4353, pp 59--73, \doi{10.1007/11965893\_5},
  \urlprefix\url{https://doi.org/10.1007/11965893\_5}

\bibitem[{Nash(1950)}]{Nash50}
Nash JF (1950) Equilibrium points in $n$-person games. Proc of the National
  Academy of Sciences 36:48--49

\bibitem[{Natarajan et~al.(2013)Natarajan, Dhillon, Ravikumar, and
  Tewari}]{DBLP:conf/nips/NatarajanDRT13}
Natarajan N, Dhillon IS, Ravikumar P, Tewari A (2013) Learning with noisy
  labels. In: Burges CJC, Bottou L, Ghahramani Z, Weinberger KQ (eds) Advances
  in Neural Information Processing Systems 26: 27th Annual Conference on Neural
  Information Processing Systems 2013. Proceedings of a meeting held December
  5-8, 2013, Lake Tahoe, Nevada, United States, pp 1196--1204,
  \urlprefix\url{https://proceedings.neurips.cc/paper/2013/hash/3871bd64012152bfb53fdf04b401193f-Abstract.html}

\bibitem[{Neumeier(2015)}]{83820}
Neumeier M (2015) The brand flip : why customers now run companies--and how to
  profit from it. New Riders,, San Francisco :

\bibitem[{Nget et~al.(2017)Nget, Cao, and Yoshikawa}]{DBLP:conf/sigir/NgetCY17}
Nget R, Cao Y, Yoshikawa M (2017) How to balance privacy and money through
  pricing mechanism in personal data market. In: Degenhardt J, Kallumadi S,
  de~Rijke M, Si L, Trotman A, Xu Y (eds) Proceedings of the {SIGIR} 2017
  Workshop On eCommerce co-located with the 40th International {ACM} {SIGIR}
  Conference on Research and Development in Information Retrieval, eCOM@SIGIR
  2017, Tokyo, Japan, August 11, 2017, CEUR-WS.org, {CEUR} Workshop
  Proceedings, vol 2311,
  \urlprefix\url{http://ceur-ws.org/Vol-2311/paper\_15.pdf}

\bibitem[{Niu et~al.(2018)Niu, Zheng, Wu, Tang, Gao, and
  Chen}]{DBLP:conf/kdd/NiuZWTGC18}
Niu C, Zheng Z, Wu F, Tang S, Gao X, Chen G (2018) Unlocking the value of
  privacy: Trading aggregate statistics over private correlated data. In: Guo
  Y, Farooq F (eds) Proceedings of the 24th {ACM} {SIGKDD} International
  Conference on Knowledge Discovery {\&} Data Mining, {KDD} 2018, London, UK,
  August 19-23, 2018, {ACM}, pp 2031--2040, \doi{10.1145/3219819.3220013},
  \urlprefix\url{https://doi.org/10.1145/3219819.3220013}

\bibitem[{Ohrimenko et~al.(2019)Ohrimenko, Tople, and
  Tschiatschek}]{DBLP:journals/corr/abs-1911-09052}
Ohrimenko O, Tople S, Tschiatschek S (2019) Collaborative machine learning
  markets with data-replication-robust payments. CoRR abs/1911.09052,
  \urlprefix\url{http://arxiv.org/abs/1911.09052}, \eprint{1911.09052}

\bibitem[{Pei(2020)}]{pei2020survey}
Pei J (2020) A survey on data pricing: from economics to data science. IEEE
  Transactions on Knowledge and Data Engineering PP:1--1,
  \doi{10.1109/TKDE.2020.3045927}

\bibitem[{Radanovic and Faltings(2016)}]{radanovic2016learning}
Radanovic G, Faltings B (2016) Learning to scale payments in crowdsourcing with
  properboost. In: Ghosh A, Lease M (eds) Proceedings of the Fourth {AAAI}
  Conference on Human Computation and Crowdsourcing, {HCOMP} 2016, 30 October -
  3 November, 2016, Austin, Texas, {USA}, {AAAI} Press, pp 179--188,
  \urlprefix\url{http://aaai.org/ocs/index.php/HCOMP/HCOMP16/paper/view/14033}

\bibitem[{Radanovic et~al.(2016)Radanovic, Faltings, and
  Jurca}]{radanovic2016incentives}
Radanovic G, Faltings B, Jurca R (2016) Incentives for effort in crowdsourcing
  using the peer truth serum. {ACM} Trans Intell Syst Technol 7(4):48:1--48:28,
  \doi{10.1145/2856102}, \urlprefix\url{https://doi.org/10.1145/2856102}

\bibitem[{Rauhut(2010)}]{rauhut2010compressive}
Rauhut H (2010) Compressive sensing and structured random matrices. Theoretical
  foundations and numerical methods for sparse recovery 9:1--92

\bibitem[{Richardson et~al.(2019)Richardson, Filos{-}Ratsikas, and
  Faltings}]{richardson2019rewarding}
Richardson A, Filos{-}Ratsikas A, Faltings B (2019) Rewarding high-quality data
  via influence functions. CoRR abs/1908.11598, \eprint{1908.11598}

\bibitem[{Richardson et~al.(2020)Richardson, Filos{-}Ratsikas, and
  Faltings}]{DBLP:series/lncs/RichardsonFF20}
Richardson A, Filos{-}Ratsikas A, Faltings B (2020) Budget-bounded incentives
  for federated learning. In: Yang Q, Fan L, Yu H (eds) Federated Learning -
  Privacy and Incentive, Lecture Notes in Computer Science, vol 12500,
  Springer, pp 176--188, \doi{10.1007/978-3-030-63076-8\_13},
  \urlprefix\url{https://doi.org/10.1007/978-3-030-63076-8\_13}

\bibitem[{Rousseeuw(1987)}]{rousseeuw1987silhouettes}
Rousseeuw PJ (1987) Silhouettes: a graphical aid to the interpretation and
  validation of cluster analysis. Journal of computational and applied
  mathematics 20:53--65

\bibitem[{Schomm et~al.(2013)Schomm, Stahl, and
  Vossen}]{schomm2013marketplaces}
Schomm F, Stahl F, Vossen G (2013) Marketplaces for data: an initial survey.
  ACM SIGMOD Record 42(1):15--26

\bibitem[{Shah and Zhou(2015)}]{shah2016double}
Shah NB, Zhou D (2015) Double or nothing: Multiplicative incentive mechanisms
  for crowdsourcing. In: Cortes C, Lawrence ND, Lee DD, Sugiyama M, Garnett R
  (eds) Advances in Neural Information Processing Systems 28: Annual Conference
  on Neural Information Processing Systems 2015, December 7-12, 2015, Montreal,
  Quebec, Canada, pp 1--9,
  \urlprefix\url{https://proceedings.neurips.cc/paper/2015/hash/c81e728d9d4c2f636f067f89cc14862c-Abstract.html}

\bibitem[{Shah and Zhou(2016)}]{DBLP:conf/icml/ShahZ16}
Shah NB, Zhou D (2016) No oops, you won't do it again: Mechanisms for
  self-correction in crowdsourcing. In: Balcan M, Weinberger KQ (eds)
  Proceedings of the 33nd International Conference on Machine Learning, {ICML}
  2016, New York City, NY, USA, June 19-24, 2016, JMLR.org, {JMLR} Workshop and
  Conference Proceedings, vol~48, pp 1--10,
  \urlprefix\url{http://proceedings.mlr.press/v48/shaha16.html}

\bibitem[{Shah et~al.(2015)Shah, Zhou, and Peres}]{DBLP:conf/icml/ShahZP15}
Shah NB, Zhou D, Peres Y (2015) Approval voting and incentives in
  crowdsourcing. In: Bach FR, Blei DM (eds) Proceedings of the 32nd
  International Conference on Machine Learning, {ICML} 2015, Lille, France,
  6-11 July 2015, JMLR.org, {JMLR} Workshop and Conference Proceedings, vol~37,
  pp 10--19, \urlprefix\url{http://proceedings.mlr.press/v37/shaha15.html}

\bibitem[{Shapley(1953)}]{shapley1953value}
Shapley LS (1953) A value for n-person games. Contributions to the Theory of
  Games 2:307--317

\bibitem[{Shnayder et~al.(2016{\natexlab{a}})Shnayder, Agarwal, Frongillo, and
  Parkes}]{DBLP:conf/sigecom/ShnayderAFP16}
Shnayder V, Agarwal A, Frongillo RM, Parkes DC (2016{\natexlab{a}}) Informed
  truthfulness in multi-task peer prediction. In: Conitzer V, Bergemann D, Chen
  Y (eds) Proceedings of the 2016 {ACM} Conference on Economics and
  Computation, {EC} '16, Maastricht, The Netherlands, July 24-28, 2016, {ACM},
  pp 179--196, \doi{10.1145/2940716.2940790},
  \urlprefix\url{https://doi.org/10.1145/2940716.2940790}

\bibitem[{Shnayder et~al.(2016{\natexlab{b}})Shnayder, Frongillo, and
  Parkes}]{DBLP:conf/ijcai/ShnayderFP16}
Shnayder V, Frongillo RM, Parkes DC (2016{\natexlab{b}}) Measuring performance
  of peer prediction mechanisms using replicator dynamics. In: Kambhampati S
  (ed) Proceedings of the Twenty-Fifth International Joint Conference on
  Artificial Intelligence, {IJCAI} 2016, New York, NY, USA, 9-15 July 2016,
  {IJCAI/AAAI} Press, pp 2611--2617,
  \urlprefix\url{http://www.ijcai.org/Abstract/16/371}

\bibitem[{Sim et~al.(2020)Sim, Zhang, Chan, and Low}]{sim2020collaborative}
Sim RHL, Zhang Y, Chan MC, Low BKH (2020) Collaborative machine learning with
  incentive-aware model rewards. In: Proceedings of the 37th International
  Conference on Machine Learning, {ICML} 2020, 13-18 July 2020, Virtual Event,
  {PMLR}, Proceedings of Machine Learning Research, vol 119, pp 8927--8936,
  \urlprefix\url{http://proceedings.mlr.press/v119/sim20a.html}

\bibitem[{Singer(2010)}]{DBLP:conf/focs/Singer10}
Singer Y (2010) Budget feasible mechanisms. In: 51th Annual {IEEE} Symposium on
  Foundations of Computer Science, {FOCS} 2010, October 23-26, 2010, Las Vegas,
  Nevada, {USA}, {IEEE} Computer Society, pp 765--774,
  \doi{10.1109/FOCS.2010.78},
  \urlprefix\url{https://doi.org/10.1109/FOCS.2010.78}

\bibitem[{Snowflake(2021)}]{SNOW}
Snowflake (2021) Snowflake data marketplace.
  \url{https://www.snowflake.com/data-marketplace/}, accessed: 2021-05-09

\bibitem[{Stahl and Vossen(2016)}]{DBLP:conf/aciids/StahlV16}
Stahl F, Vossen G (2016) Data quality scores for pricing on data marketplaces.
  In: Nguyen NT, Trawinski B, Fujita H, Hong T (eds) Intelligent Information
  and Database Systems - 8th Asian Conference, {ACIIDS} 2016, Da Nang, Vietnam,
  March 14-16, 2016, Proceedings, Part {I}, Springer, Lecture Notes in Computer
  Science, vol 9621, pp 215--224, \doi{10.1007/978-3-662-49381-6\_21},
  \urlprefix\url{https://doi.org/10.1007/978-3-662-49381-6\_21}

\bibitem[{Tang et~al.(2013)Tang, Wu, Bao, Bressan, and
  Valduriez}]{DBLP:conf/dexa/TangWBBV13}
Tang R, Wu H, Bao Z, Bressan S, Valduriez P (2013) The price is right - models
  and algorithms for pricing data. In: Decker H, Lhotsk{\'{a}} L, Link S, Basl
  J, Tjoa AM (eds) Database and Expert Systems Applications - 24th
  International Conference, {DEXA} 2013, Prague, Czech Republic, August 26-29,
  2013. Proceedings, Part {II}, Springer, Lecture Notes in Computer Science,
  vol 8056, pp 380--394, \doi{10.1007/978-3-642-40173-2\_31},
  \urlprefix\url{https://doi.org/10.1007/978-3-642-40173-2\_31}

\bibitem[{Tang et~al.(2014)Tang, Amarilli, Senellart, and
  Bressan}]{DBLP:conf/dexa/TangASB14}
Tang R, Amarilli A, Senellart P, Bressan S (2014) Get a sample for a discount -
  sampling-based {XML} data pricing. In: Decker H, Lhotsk{\'{a}} L, Link S,
  Spies M, Wagner RR (eds) Database and Expert Systems Applications - 25th
  International Conference, {DEXA} 2014, Munich, Germany, September 1-4, 2014.
  Proceedings, Part {I}, Springer, Lecture Notes in Computer Science, vol 8644,
  pp 20--34, \doi{10.1007/978-3-319-10073-9\_3},
  \urlprefix\url{https://doi.org/10.1007/978-3-319-10073-9\_3}

\bibitem[{Tram{\`{e}}r et~al.(2016)Tram{\`{e}}r, Zhang, Juels, Reiter, and
  Ristenpart}]{DBLP:conf/uss/TramerZJRR16}
Tram{\`{e}}r F, Zhang F, Juels A, Reiter MK, Ristenpart T (2016) Stealing
  machine learning models via prediction apis. In: Holz T, Savage S (eds) 25th
  {USENIX} Security Symposium, {USENIX} Security 16, Austin, TX, USA, August
  10-12, 2016, {USENIX} Association, pp 601--618,
  \urlprefix\url{https://www.usenix.org/conference/usenixsecurity16/technical-sessions/presentation/tramer}

\bibitem[{Upadhyaya et~al.(2016)Upadhyaya, Balazinska, and
  Suciu}]{DBLP:journals/pvldb/UpadhyayaBS16}
Upadhyaya P, Balazinska M, Suciu D (2016) Price-optimal querying with data
  apis. Proc {VLDB} Endow 9(14):1695--1706, \doi{10.14778/3007328.3007335},
  \urlprefix\url{http://www.vldb.org/pvldb/vol9/p1695-upadhyaya.pdf}

\bibitem[{Vaughan(2017)}]{vaughan2017making}
Vaughan JW (2017) Making better use of the crowd: How crowdsourcing can advance
  machine learning research. J Mach Learn Res 18:193:1--193:46,
  \urlprefix\url{http://jmlr.org/papers/v18/17-234.html}

\bibitem[{Wang et~al.(2020)Wang, Rausch, Zhang, Jia, and
  Song}]{DBLP:series/lncs/0013RZJS20}
Wang T, Rausch J, Zhang C, Jia R, Song D (2020) A principled approach to data
  valuation for federated learning. In: Yang Q, Fan L, Yu H (eds) Federated
  Learning - Privacy and Incentive, Lecture Notes in Computer Science, vol
  12500, Springer, pp 153--167, \doi{10.1007/978-3-030-63076-8\_11},
  \urlprefix\url{https://doi.org/10.1007/978-3-030-63076-8\_11}

\bibitem[{Yan and Procaccia(2021)}]{DBLP:conf/aaai/YanP21}
Yan T, Procaccia AD (2021) If you like shapley then you'll love the core. In:
  Thirty-Fifth {AAAI} Conference on Artificial Intelligence, {AAAI} 2021,
  Thirty-Third Conference on Innovative Applications of Artificial
  Intelligence, {IAAI} 2021, The Eleventh Symposium on Educational Advances in
  Artificial Intelligence, {EAAI} 2021, Virtual Event, February 2-9, 2021,
  {AAAI} Press, pp 5751--5759,
  \urlprefix\url{https://ojs.aaai.org/index.php/AAAI/article/view/16721}

\bibitem[{Yang et~al.(2012)Yang, Xue, Fang, and Tang}]{yang2012crowdsourcing}
Yang D, Xue G, Fang X, Tang J (2012) Crowdsourcing to smartphones: incentive
  mechanism design for mobile phone sensing. In: Akan {\"{O}}B, Ekici E, Qiu L,
  Snoeren AC (eds) The 18th Annual International Conference on Mobile Computing
  and Networking, Mobicom'12, Istanbul, Turkey, August 22-26, 2012, {ACM}, pp
  173--184, \doi{10.1145/2348543.2348567},
  \urlprefix\url{https://doi.org/10.1145/2348543.2348567}

\bibitem[{Yoon et~al.(2020)Yoon, Arik, and Pfister}]{DBLP:conf/icml/YoonAP20}
Yoon J, Arik S{\"{O}}, Pfister T (2020) Data valuation using reinforcement
  learning. In: Proceedings of the 37th International Conference on Machine
  Learning, {ICML} 2020, 13-18 July 2020, Virtual Event, {PMLR}, Proceedings of
  Machine Learning Research, vol 119, pp 10842--10851,
  \urlprefix\url{http://proceedings.mlr.press/v119/yoon20a.html}

\bibitem[{Yu and Zhang(2017)}]{DBLP:journals/candie/YuZ17}
Yu H, Zhang M (2017) Data pricing strategy based on data quality. Comput Ind
  Eng 112:1--10, \doi{10.1016/j.cie.2017.08.008},
  \urlprefix\url{https://doi.org/10.1016/j.cie.2017.08.008}

\bibitem[{Yu et~al.(2020{\natexlab{a}})Yu, Liu, Liu, Chen, Cong, Weng, Niyato,
  and Yang}]{DBLP:journals/expert/YuLLCCWNY20}
Yu H, Liu Z, Liu Y, Chen T, Cong M, Weng X, Niyato D, Yang Q
  (2020{\natexlab{a}}) A sustainable incentive scheme for federated learning.
  {IEEE} Intell Syst 35(4):58--69, \doi{10.1109/MIS.2020.2987774},
  \urlprefix\url{https://doi.org/10.1109/MIS.2020.2987774}

\bibitem[{Yu et~al.(2020{\natexlab{b}})Yu, Yang, Zhang, Tsai, Ho, and
  Jin}]{DBLP:conf/ndss/YuYZTHJ20}
Yu H, Yang K, Zhang T, Tsai Y, Ho T, Jin Y (2020{\natexlab{b}}) Cloudleak:
  Large-scale deep learning models stealing through adversarial examples. In:
  27th Annual Network and Distributed System Security Symposium, {NDSS} 2020,
  San Diego, California, USA, February 23-26, 2020, The Internet Society,
  \urlprefix\url{https://www.ndss-symposium.org/ndss-paper/cloudleak-large-scale-deep-learning-models-stealing-through-adversarial-examples/}

\bibitem[{Zhang and Beltran(2020)}]{ZhangBletran20}
Zhang M, Beltran F (2020) A survey of data pricing methods. SSRN

\bibitem[{Zhang et~al.(2020)Zhang, Beltr{\'{a}}n, and
  Liu}]{DBLP:conf/uai/ZhangBL20}
Zhang M, Beltr{\'{a}}n F, Liu J (2020) Selling data at an auction under privacy
  constraints. In: Adams RP, Gogate V (eds) Proceedings of the Thirty-Sixth
  Conference on Uncertainty in Artificial Intelligence, {UAI} 2020, virtual
  online, August 3-6, 2020, {AUAI} Press, Proceedings of Machine Learning
  Research, vol 124, pp 669--678,
  \urlprefix\url{http://proceedings.mlr.press/v124/zhang20b.html}

\bibitem[{Zhang et~al.(2016)Zhang, Yang, Sun, Liu, Tang, Xing, and
  Mao}]{DBLP:journals/comsur/ZhangYSLTXM16}
Zhang X, Yang Z, Sun W, Liu Y, Tang S, Xing K, Mao X (2016) Incentives for
  mobile crowd sensing: {A} survey. {IEEE} Commun Surv Tutorials 18(1):54--67,
  \doi{10.1109/COMST.2015.2415528},
  \urlprefix\url{https://doi.org/10.1109/COMST.2015.2415528}

\bibitem[{Zhou and Zheng(2009)}]{zhou2009trust}
Zhou X, Zheng H (2009) Trust: A general framework for truthful double spectrum
  auctions. In: IEEE INFOCOM 2009, IEEE, pp 999--1007

\bibitem[{Zhou et~al.(2014)Zhou, Porwal, Zhang, Ngo, Nguyen, R{\'{e}}, and
  Govindaraju}]{DBLP:conf/nips/ZhouPZNNRG14}
Zhou Y, Porwal U, Zhang C, Ngo HQ, Nguyen L, R{\'{e}} C, Govindaraju V (2014)
  Parallel feature selection inspired by group testing. In: Ghahramani Z,
  Welling M, Cortes C, Lawrence ND, Weinberger KQ (eds) Advances in Neural
  Information Processing Systems 27: Annual Conference on Neural Information
  Processing Systems 2014, December 8-13 2014, Montreal, Quebec, Canada, pp
  3554--3562,
  \urlprefix\url{https://proceedings.neurips.cc/paper/2014/hash/fb8feff253bb6c834deb61ec76baa893-Abstract.html}

\end{thebibliography}

\end{sloppy}
\end{document}